\documentclass{elsarticle}

\usepackage{epsfig}
\usepackage{epsfig,latexsym, amssymb, amsmath}
\usepackage{algorithmic, algorithm}
\usepackage{lscape}
\newdefinition{example}{Example}
\begin{document}
\begin{frontmatter}


\title{Evaluating and Improving Modern Variable and Revision Ordering Strategies in CSPs}

\author[Aegean]{Thanasis Balafoutis\corref{cor}}
\ead{abalafoutis@aegean.gr}

\author[Aegean]{Kostas Stergiou}
\ead{konsterg@aegean.gr}

\address[Aegean]{Department of Information and Communication Systems
Engineering, University of the Aegean, Samos, Greece}

\cortext[cor]{Department of Information \& Communication Systems
Engineering, University of the Aegean, Karlovassi, Samos, 83200, Greece. Tel: +30 6937 309079, Fax: +30 22730 82229}

\begin{abstract}
A key factor that can dramatically reduce the search space during
constraint 
solving is the criterion under which the variable to be instantiated
next is selected. For this purpose numerous heuristics have been
proposed. Some of the best of such heuristics exploit information
about failures gathered throughout search and recorded in the form
of constraint weights, while others measure the importance of
variable assignments in reducing the search space. In this work we
experimentally evaluate the most recent and powerful variable
ordering heuristics, and new variants of them, over a wide range
of benchmarks. Results demonstrate that heuristics based on
failures are in general more efficient. Based on this, we then
derive new revision ordering heuristics that exploit recorded
failures to efficiently order the propagation list when arc
consistency is maintained during search. Interestingly, in
addition to reducing the number of constraint checks and list
operations, these heuristics are also able to cut down the size of
the explored search tree.

\end{abstract}


\end{frontmatter}

\section{Introduction}

Constraint programming is a powerful technique for solving
combinatorial search problems that draws on a wide range of
methods from artificial intelligence and computer science. The
basic idea in constraint programming is that the user states the
constraints and a general purpose constraint solver is used to
solve the resulting constraint satisfaction problem. Since constraints are relations,
a Constraint Satisfaction Problem (CSP) states which relations hold among the given decision variables. CSPs can be solved either
systematically, as with backtracking, or using forms of local search which may be
incomplete. When solving a CSP using backtracking search, a
sequence of decisions must be made as to which variable to
instantiate next. These decisions are referred to as the variable
ordering decisions. It has been shown that for many problems the
choice of variable ordering can have a dramatic effect on the
performance of the backtracking algorithm with huge variances even
on a single instance \cite{gentMac96,vanbeekBook06}.

A variable ordering can be either static, where the ordering is
fixed and determined prior to search, or dynamic, where the
ordering is determined as the search proceeds. Dynamic variable
orderings are considerably more efficient and have thus received much attention in the literature. One
common dynamic variable ordering strategy, known as
``fail-first'', is to select as the next variable the one likely
to fail as quickly as possible.

Recent years have seen the emergence of numerous modern heuristics
for choosing variables during CSP search. The so called
conflict-driven heuristics exploit information about failures
gathered throughout search and recorded in the form of constraint
weights, while other heuristics measure the importance of variable
assignments in reducing the search space. Most of them are quite
successful and choosing the best general purpose heuristic is not
easy. All these new heuristics have been tested over a narrow set
of problems in their original papers and they have been compared
mainly with older heuristics. Hence, there is no comprehensive view of
the relative strengths and weaknesses of these heuristics.

This paper is an improvement to that published previously in \cite{balstergRCRA08}.
A first aim of the present work is to experimentally evaluate the
performance of the most recent and powerful heuristics over a wide
range of benchmarks, in order to reveal their strengths and
weaknesses. Results demonstrate that conflict-driven heuristics
such as the well known \emph{dom/wdeg} heuristic \cite{bhls04} are in general
faster and more robust than other heuristics. Based on these
results, as a second contribution, we have tried to improve the
behavior of the \emph{dom/wdeg} heuristic resulting in interesting
additions to the family of conflict-driven heuristics.

We also investigate new ways to exploit failures in order to speed
up constraint solving. To be precise, we investigate the
interaction between conflict-driven variable ordering heuristics
and revision list ordering heuristics and propose new efficient
revision ordering heuristics. Constraint solvers that maintain a
local consistency (e.g. Maintaining Arc Consistency, MAC-based solvers) employ a {\em revision
list} of variables, constraints, or (hyper)arcs 
(depending on the implementation), to propagate the effects of
variable assignments. It has been shown that the order in which
the elements of the list are selected for revision affects the
overall cost of the search. Hence, a number of revision ordering
heuristics have been proposed and evaluated
\cite{wallace92,boussem04,schulte08}. In general, variable
ordering and revision ordering heuristics have different tasks to
perform when used by a search algorithm such as MAC. Prior to the
emergence of conflict-driven variable ordering heuristics it was
not possible to achieve an interaction with each other, i.e. the
order in which the revision list was organized during propagation
could not affect the decision of which variable to select next
(and vice versa). The contribution of revision ordering heuristics
to the solver's efficiency was limited to the reduction of list
operations and constraint checks.

We demonstrate that when a conflict-driven variable ordering
heuristic like \emph{dom/wdeg} is used, then there are cases where
the order in the elements of the list are revised can
affect the variable selection.
Inspired by this, a third contribution of this paper is to propose
new, conflict-driven, heuristics for ordering the revision list.
We show that these heuristics can not only reduce the numbers of
constraints checks and list operations, but also cut down the size
of the explored search tree. Results from various benchmarks
demonstrate that some of the proposed heuristics can boost the
performance of the \emph{dom/wdeg} heuristic up to 5 times.
Interestingly, we also show that some of the new variants of
\emph{dom/wdeg} that we propose are much less amenable to the
revision ordering than \emph{dom/wdeg}.

The main contributions of this paper can be summarized as follows:

\begin{itemize}
  \item We give experimental results from a detailed comparison of modern variable ordering
heuristics in a wide range of academic, random and real world
problems. These experiments demonstrate that \emph{dom/wdeg} and
its variants can be considered the most efficient and robust among
the heuristics compared.


  \item Based on our observation concerning the interaction
  between conflict-driven variable ordering heuristics and
  revision ordering heuristics, we extend the use of failures
  discovered during search
  to devise new and efficient revision ordering heuristics. These heuristics
  can increase the efficiency of the solver by not only reducing list operation but also by cutting
  down the size of the explored search tree.

  \item We show that certain variants of \emph{dom/wdeg}
  are less amenable to changes in the revision ordering than
  \emph{dom/wdeg} and therefore can be more robust.
\end{itemize}

The rest of the paper is organized as follows.
Section~\ref{sec:background} gives the necessary background
material. In Section~\ref{sec:heuristics} we give an overview of
existing variable ordering heuristics. In
Section~\ref{sec:experiments} we present and discuss the
experimental results from a wide variety of real world, academic and
random problems. In Section~\ref{sec:revision} after a short
summary on the existing revision ordering heuristics for
constraint propagation, we propose a set of new revision ordering
heuristics based on constraint weights. We then give experimental
results comparing these heuristics with existing ones.
Section~\ref{sec:revision} concludes with a discussion and some
experimental results on the dependency between conflict-driven
variable ordering heuristics and revision orderings. Finally, conclusions are presented in
Section~\ref{sec:conclusion}.

\section{Background}

\label{sec:background}

A \emph{Constraint Satisfaction Problem} (CSP) is a tuple
(\emph{X, D, C}), where \emph{X} is a set containing \emph{n}
variables \{\emph{$x_1, x_2,..., x_n$}\}; \emph{D} is a set of
domains \{\emph{$D(x_1)$, $D(x_2)$,..., $D(x_n)$}\} for those
variables, with each $D(x_i)$ consisting of the possible values
which $x_i$ may take; and \emph{C} is a set of $e$ constraints
\{\emph{$c_1, c_2,..., c_e$}\} between variables in subsets of
\emph{X}. Each $c_i \in C$ expresses a relation defining which
variable assignment combinations are allowed for the variables in
the scope of the constraint, \emph{vars($c_i$)}. Two variables are
said to be \emph{neighbors} if they share a constraint. The
\emph{arity} of a constraint is the number of variables in the
scope of the constraint. The \emph{degree} of a variable $x_i$,
denoted by $\Gamma(x_i)$, is the number of constraints in which
$x_i$ participates. A binary constraint between variables $x_i$
and $x_j$ will be denoted by $c_{ij}$.

A partial assignment is a set of tuple pairs, each tuple
consisting of an instantiated variable and the value that is
assigned to it in the current search node. A full assignment is
one containing all \emph{n} variables. A solution to a CSP is a
full assignment such that no constraint is violated.

In binary CSPs any constraint $c_{ij}$ defines two directed arcs
($x_i$,$x_j$) and ($x_j$,$x_i$). 
A directed constraint ($x_i$,$x_j$) is \emph{arc consistent} (AC) iff for every
value $a \in D(x_i)$ there exists at least one value $b\in D(x_j)$
such that the pair ($a$,$b$) satisfies $c_{ij}$. In this case we
say that $b$ is a {\em support} of $a$ on the directed constraint ($x_i$,$x_j$).
Accordingly, $a$ is a support of $b$ on the directed constraint ($x_j$,$x_i$). A
problem is AC iff there are no empty domains and all arcs are AC.
Enforcing AC on a problem results in the removal of all
non-supported values from the domains of the variables. The
definition of arc consistency for non-binary constraints, usually
called \emph{generalized arc consistency} (GAC), is a direct
extension of the definition of AC. A non-binary constraint $c$,
with $vars(c)$=\{\emph{$x_1, x_2,..., x_k$}\}, is GAC iff for
every variable $x_i \in vars(c)$ and every value $a \in D(x_i)$
there exists a tuple $\tau$ that satisfies $c$ and includes the
assignment of $a$ to $x_i$ \cite{mh86,mackb77}. In this case
$\tau$ is a support of $a$ on constraint $c$. A problem is GAC iff
all constraints are GAC. In the rest of the paper we will assume
that (G)AC is the propagation method applied to all constraints.

Many consistency properties and corresponding propagation algorithms
stron-ger than AC and GAC have been proposed in the literature. One
of the most studied is singleton (G)AC which, as we will explain in
the following section, has also been used to guide the selection
process for a certain variable ordering heuristic.
A variable $x_i$ is {\em singleton generalized arc consistent}
(SGAC) iff for each value $a_i\in D(x_i)$, after assigning $a_i$ to
$x_i$ and applying GAC in the problem, there is no empty domain \cite{db01}.

A \emph{support check} (consistency check) is a test to find out
if a tuple supports a given value. In the case of binary CSPs a
support check simply verifies if two values support each other or
not. The \emph{revision} of a variable-constraint pair $(c,x_i)$,
with $x_i \in vars(c)$, verifies if all values in $D(x_i)$ have
support on $c$. In the binary case the revision of an arc
($x_i$,$x_j$) verifies if all values in \emph{$D(x_i)$} have
supports in $D(x_j)$. We say that a revision is \emph{fruitful} if
it deletes at least one value, while it is \emph{redundant} if it
achieves no pruning. A \emph{DWO-revision} is one that causes a
domain wipeout (\emph{DWO}). That is, it removes the last
remaining value(s) from a domain.

Complete search algorithms for CSPs are typically based on
backtracking depth-first search where branching decisions (i.e.
variable assignments) are interleaved with constraint propagation.
The search algorithm used in the experiments presented is known as
MGAC (maintaining generalized arc consistency) or MAC in the case
of binary problems \cite{sabin94,bessiere96}. This algorithm can
be implemented using either a \emph{d-way} or a \emph{2-way}
branching scheme. The former works as follows.
Initially, the whole problem should be made GAC before starting search.
After the first variable $x$ with domain $D(x)=\{a_1, a_2,..., a_d\}$ is selected, $d$
recursive calls are made. In the first call value $a_1$ is
assigned to $x$ and the problem is made GAC, i.e. all values which
are not GAC given the assignment of $a_1$ to $x$ are removed. If
this call fails (i.e. no solution is found), the value $a_1$ is removed from the domain of $x$
and the problem is made again GAC. Then a second recursive
call under the assignment of $a_2$ to $x$ is made, and so on. The
problem has no solution if all $d$ calls fail.
In \emph{2-way} branching, after a variable $x$ and a value $a_i
\in D(x)$ are selected, two recursive calls are made. In the first
call $a_i$ is assigned to $x$, or in other words the constraint
$x$=$a_i$ is added to the problem, and GAC is applied. In the
second call the constraint $x\neq a_i$ is added to the problem and
GAC is applied. The problem has no solution if neither recursive
call finds a solution. The main difference of these branching schemes is that
in \emph{2-way} branching, after a failed choice of a variable assignment ($x$,$a_i$) the algorithm can choose a new assignment for any variable (not only $x$). In \emph{d-way} branching the algorithm has to choose the next available value for variable $x$.

\section{Overview of variable ordering heuristics}

\label{sec:heuristics}

The order in which variables are assigned by a backtracking search
algorithm has been understood for a long time to be of primary
importance. The first category of heuristics used for ordering
variables was based on the initial structure of the network. These
are called static or fixed variable ordering heuristics (SVOs) as
they simply replace the lexicographic ordering by something more
appropriate to the structure of the network before starting search.
Examples of such heuristics are \emph{min width} which chooses an
ordering that minimizes the width of the constraint network
\cite{freuder82}, \emph{min bandwidth} which minimizes the
bandwidth of the constraint graph
\cite{zabih90}, and \emph{max degree} (\emph{deg}), 
where variables are ordered according to the initial size of their neighborhood \cite{dechter89}.

A second category of heuristics includes dynamic variable ordering
heuristics (DVOs) which take into account information about the
current state of the problem at each point in the search. The first
well known dynamic heuristic, introduced by Haralick and Elliott,
was \emph{dom} \cite{haral80}. This heuristic chooses the variable
with the smallest remaining domain. The dynamic variation of
\emph{deg}, called \emph{ddeg} selects the variable with largest
dynamic degree. That is, for binary CSPs, the variable that is constrained with the
largest number of unassigned variables. By combining \emph{dom}
and \emph{deg} (or \emph{ddeg}), the heuristics called
\emph{dom/deg} and \emph{dom/ddeg} \cite{bessiere96,smith98} were
derived. These heuristics select the variable that minimizes the
ratio of current domain size to static degree (dynamic degree) and
can significantly improve the search performance.

When using variable ordering heuristics, it is a common phenomenon
that ties can occur. A tie is a situation where a number of
variables are considered equivalent by a heuristic. Especially at
the beginning of search, where it is more likely that the domains
of the variables are of equal size, ties are frequently noticed. A
common tie breaker for the \emph{dom} heuristic is \emph{lexico},
(\emph{dom+lexico} composed heuristic) which selects among the
variables with smallest domain size the lexicographically first.
Other known composed heuristics are \emph{dom+deg} \cite{frost95},
\emph{dom+ddeg} \cite{brelaz79,smith99} and \emph{BZ3}
\cite{smith99}.

Bessi\`ere et al. \cite{beschmsais01}, have proposed a general
formulation of DVOs which integrates in the selection function a
measure of the constrainedness of the given variable. These
heuristics (denoted as \emph{mDVO}) take into account the
variable's neighborhood and they can be considered as neighborhood
generalizations of the $dom$ and $dom/ddeg$ heuristics. For
instance, the selection function for variable $X_i$ is described
as follows:

\begin{equation}
H_a^{\circledcirc}(x_i) = \frac{\sum_{{x_j}\in\Gamma(x_i)} (\alpha(x_i) \circledcirc \alpha(x_j))}{|\Gamma(x_i)|^2}
\end{equation}

where $\Gamma(x_i)$ is the set of variables that share a constraint with $x_i$
and $\alpha(x_i)$ can be any simple syntactical property of the
variable such as $|D(x_i)|$ or $\frac{|D(x_i)|}{|\Gamma(x_i)|}$
and $\circledcirc \in \{+,\times\}$. Neighborhood based heuristics
have shown to be quite promising.

Boussemart et al. \cite{bhls04}, inspired from SAT (satisfiability testing)
solvers like Chaff \cite{mosk01}, proposed conflict-driven
variable ordering heuristics. In these heuristics, every time a
constraint causes a failure (i.e. a domain wipeout) during search,
its weight is incremented by one. Each variable has a
\emph{weighted degree}, which is the sum of the weights over all
constraints in which this variable participates. The weighted
degree heuristic (\emph{wdeg}) selects the variable with the
largest weighted degree. The current domain of the variable can
also be incorporated to give the domain-over-weighted-degree
heuristic (\emph{dom/wdeg}) which selects the variable with
minimum ratio between current domain size and weighted degree.
Both of these heuristics (especially \emph{dom/wdeg}) have been
shown to be very effective on a wide range of problems.

Grimes and Wallace \cite{grim07,grim08} proposed alternative
conflict-driven heuristics that consider value deletions as the
basic propagation events associated with constraint weights. That
is, the weight of a constraint is incremented each time the
constraint causes one or more value deletions. They also used a
sampling technique called \emph{random probing} where several short runs of the search algorithm are made to initialize the constraint weights prior to the final run. Using this method \emph{global contention}, i.e. contention that holds across the entire search space, can be uncovered.

Inspired by integer programming, Refalo introduced an
\emph{impact} measure with the aim of detecting choices which
result in the strongest search space reduction \cite{refal04}. An
impact is an estimation of the importance of a value assignment
for reducing the search space. Refalo proposes to characterize the
impact of a decision by computing the Cartesian product of the
domains before and after the considered decision. The impacts of
assignments for every value can be approximated by the use of
averaged values at the current level of observation. If $K$ is the
index set of impacts observed so far for assignment $x_i =
\alpha$, $\overline{I}$ is the averaged impact:

\begin{equation}
\overline{I}(x_i=\alpha)=\frac{\displaystyle\sum_{k \in K} I^k(x_i=\alpha)}{|K|}
\end{equation}

where $I^k$ is the observed value impact for any $k \in K$.

The impact of a variable $x_i$ can be computed by the following equation:

\begin{equation}
I(x_i)=\displaystyle\sum_{\alpha \in D(x_i)} 1-\overline{I}(x_i=\alpha)
\end{equation}

An interesting extension of the above heuristic is the use of
``node impacts'' to break ties in a subset of variables that have
equivalent impacts. Node impacts are the accurate impact values
which can be computed for any variable by trying all possible
assignments.

Correia and Barahona \cite{correia07} proposed variable orderings,
by integrating Singleton Consistency propagation procedures with
look-ahead heuristics. This heuristic is similar to ``node
impacts'', but instead of computing the accurate impacts, it
computes the reduction in the search space after the application
of Restricted Singleton Consistency (RSC) \cite{prosser00}, for
every value of the current variable. Although this heuristic was
firstly introduced to break ties in variables with current domain
size equal to 2, it can also be used as a tie breaker for any
other variable ordering heuristic.

Cambazard and Jussien \cite{cambaz06} went a step further by
analyzing where the reduction of the search space occurs and how
past choices are involved in this reduction. This is implemented
through the use of \emph{explanations}. An explanation consists of
a set of constraints $C'$ (a subset of the set C of the original
constraints of the problem) and a set of decisions $dc_1,...,dc_n$
taken during search.

Zanarini and Pesant \cite{zanar07} proposed \emph{constraint-centered heuristics} which
guide the exploration of the search space toward areas that are likely to contain
a high number of solutions. These heuristics are based on solution counting information
at the level of individual constraints. Although the cost of computing the solution counting
information is in general large, it has been shown that for certain widely-used global constraints, such information can be computed efficiently.


Finally, we proposed \cite{balstergSETN10} new
variants of conflict-driven heuristics. These variants differ from
\emph{wdeg} in the way they assign weights. They propose
heuristics that record the constraint that is responsible for any
value deletion during search, heuristics that give greater
importance to recent conflicts, and finally heuristics that try to
identify contentious constraints by detecting all possible
conflicts after a failure. The last heuristic, called ``fully
assigned'', increases the weights of constraints that are
responsible for a DWO by one (as $wdeg$ heuristic does) and also,
only for revision lists that lead to a DWO, increases by one the
weights of constraints that participate in fruitful revisions
(revisions that delete at least one value). Hence, this heuristic
records all variables that delete at least one value during
constraint propagation and if a DWO is detected, it increases the
weight of all these variables by one.

\section{Experiments and results}

\label{sec:experiments}

We now report results from the experimental evaluation of the
selected DVOs described above on several classes of problems.
All benchmarks are taken from C. Lecoutre's web page
(http://www.cril.univ-artois.fr/$\sim$lecoutre/research /benchmarks/),
where the reader can find additional details on how the benchmarks
are constructed. In our experiments we included both satisfiable
and unsatisfiable instances. Each selected instance involves
constraints defined either in intension or in extension. Our solver can
accept any kind of intentional constraints that are supported by the XCSP 2.1 format \cite{xcsp09}
(The XML format that were used to represent constraint networks in the last international competition
of CSP solvers).

We have tried to include a wide range of of CSP instances from
different backgrounds. Hence, we have experimented with instances from
real world applications, instances following a regular pattern and
involving a random generation, academic instances which
do not involve any random generation, random instances containing
a small structure, pure random instances and, finally, instances
which involve only Boolean variables. The selected instances
include both binary and non-binary constraints. In
Table~\ref{table:bench} we give the total number of tested
instances on each problem category. In this section we only
present results from a subset of the tried instances. In some
cases different instances within the same problem class displayed
very similar behavior with respect to their difficulty (measured
in cpu times and node visits). In such cases we only include
results from one of these instances. Also, we do not present
results from some very easy and some extremely hard instances.

\begin{table}
\caption{Problem categories that have been included in the experiments and the corresponding number of tested instances}
\centering
\begin{scriptsize}
\begin{tabular}{|c|c|}
\hline

\textbf{CSP category} & \textbf{number of instances} \\
\hline 
Real world & 80 \\ \hline
Patterned & 36 \\ \hline
Academic & 48 \\ \hline
Quasi random & 28 \\ \hline
Pure random & 36 \\ \hline
Boolean & 92 \\ \hline
\end{tabular}
\end{scriptsize}
\label{table:bench} 
\end{table}

The CSP solver\footnote{The solver is available on
request from the first author.} used in our experiments is a generic solver (in the
sense that it can handle constraints of any arity) and has been
implemented in the Java programming language. This solver
essentially implements the M(G)AC search algorithm, where (G)AC-3
is used for applying (G)AC. Although numerous other generic (G)AC
algorithms exist in the literature, especially for binary
constraints, (G)AC-3 is quite competitive despite being one of the
simplest. The solver uses d-way branching and can apply any given
restart policy. All experiments were run on an Intel dual core PC
T4200 2GHz with 3GB RAM.

Concerning the performance of our solver compared to two state-of-the-art solvers, Abscon 109 \cite{abscon08} and Choco \cite{choco08}, some preliminary results showed that all three solvers visited roughly the same amount of nodes, our solver was consistently slower than Abscon, but sometimes faster than Choco. Note that the aim of our study is to fairly compare the various variable ordering heuristics within the same solver's environment and not to build a state-of-the-art constraint solver. Although our implementation is reasonably optimized for its purposes, it lacks important aspects of state-of-the-art constraint solvers such as specialized propagators for global constraints and intricate data structures. On the other hand, we are not aware of any solver, commercial or not, that offers all of the variable ordering heuristics tested here (see Subsection~\ref{subsec:details}).

Concerning the experiments, most results were obtained using a
lexicographic value ordering, but we also evaluated the impact of
random value ordering on the relative performance of the
heuristics. We employed a geometric restart policy where the
initial number of allowed backtracks for the first run was set to
10 and at each new run the number of allowed backtracks increased
by a factor of 1.5. In addition, we evaluated the heuristics under
a different restart policy and in the absence of restarts. Since
our solver does not yet support global constraints, we have left
experiments with problems that include such constraints as
future work.

In our experiments the random probing technique is run to a fixed
failure-count cutoff C = 40, and for a fixed number of restarts R
= 50 (these are the optimal values from \cite{grim07}). After the
random probing phase has finished, search starts with the
failure-count cutoff being removed and the dom/wdeg heuristic used
based on the accumulated weights for each variable. According to
\cite{grim07}, there are two strategies one can pursue during
search. The first is to use the weights accumulated through
probing as the final weights for the constraints. The second is to
continue to increment them during search in the usual way. In our
experiments we have used the latter approach. Cpu time and nodes
for random probing are averaged values for 50 runs. For heuristics
that use probing we have measured the total cpu time and the total
number of visited nodes (from both random probing initialization
and final search). In the next tables (except
Table~\ref{table:rlfap}) we also show in parenthesis results from
the final search only (with the random probing initialization
overhead excluded).

Concerning impacts, we have approximated their values at the
initialization phase by dividing the domains of the variables into
(at maximum) four sub-domains.

As a primary parameter for the measurement of performance of the
evaluated strategies, we have used the cpu time in seconds (t). We
have also recorded the number of visited nodes (n) as this gives a
measure that is not affected by the particular implementation or by the
hardware used. In all the experiments, a time out limit has been
set to 1 hour.

In Section~\ref{subsec:details} we give some additional details
on the heuristics which we have selected for the evaluation.
In Section~\ref{subsec:rlfap} we present results from the radio link frequency
assignment problem (RLFAP). In Section~\ref{subsec:struct} we present results from
structured and patterned problems. These instances are taken from some academic
(langford), real world (driver) and patterned (graph coloring)
problems. In Section~\ref{subsec:rand} we consider instances from quasi-random
and random problems.  Experiments with non-binary constraints are presented in Section~\ref{subsec:nonbinary}.
The last experiments presented in Section~\ref{subsec:bool}
 include Boolean instances. In Section~\ref{subsec:restarts}, we study the impact of the selected restart
 policy on the evaluated heuristics, while in Section~\ref{subsec:valueOrdering} we present experiments with
 random value ordering.
 Finally in Section~\ref{subsec:general} we make a
general discussion where we summarize our results.

\subsection{Details on the evaluated heuristics}

\label{subsec:details}

For the evaluation we have selected heuristics from 5 recent
papers mentioned above. These are: i) \emph{dom/wdeg} from
Boussemart et al. \cite{bhls04}, ii) the random probing technique
and the ``alldel by \#del'' heuristic where constraint weights are
increased by the size of the domain reduction (Grimes and Wallace
\cite{grim07}), iii) Impacts and Node Impacts from Refalo
\cite{refal04}, iv) the ``RSC'' heuristic from Correia and
Barahona \cite{correia07} and, finally, v) our ``fully assigned''
heuristic \cite{balstergSETN10}.

We have also included in our experiments some combinations of the
above heuristics. For example, \emph{dom/wdeg} can be combined
with RSC (in this case RSC is used only to break ties). Random
probing can be applied to any conflict-driven heuristic, hence it
can be used with the \emph{dom/wdeg} and ``fully assigned''
heuristics. Moreover, the impact heuristic can be combined with
RSC for breaking ties.

The full list of the heuristics that we have tried in our
experiments includes 15 variations. These are the following: 1)
dom/wdeg, 2) dom/wdeg + RSC (the second heuristic is used only for
breaking ties), 3) dom/wdeg with random probing, 4) dom/wdeg with
random probing + RSC, 5) Impacts, 6) Node Impacts, 7) Impacts +
RSC, 8) alldel by \#del, 9) alldel by \#del + RSC, 10) alldel by
\#del with random probing, 11) alldel by \#del with random probing
+ RSC, 12) fully assigned, 13) fully assigned + RSC, 14) fully
assigned with random probing, and 15) fully assigned with random
probing + RSC. In all these variations the RSC heuristic is used
only for breaking ties.

\subsection{RLFAP instances}

\label{subsec:rlfap}

The Radio Link Frequency Assignment Problem (RLFAP) is the task of
assigning frequencies to a number of radio links so that a large
number of constraints are simultaneously satisfied and as few
distinct frequencies as possible are used. A number of modified
RLFAP instances have been produced from the original set of
problems. These instances have been translated into pure satisfaction
problems after removing some frequencies (denoted by f followed by a
value)\cite{cabon99}. For example, scen11-f8 corresponds to the instance scen11
for which the 8 highest frequencies have been removed.

Results from Table~\ref{table:rlfap} show that conflict-driven
heuristics (dom/wdeg, alldel and fully assigned) have the best
performance. In the final line of Table~\ref{table:rlfap} we give
the averaged values for all the instances.

\newpage

\landscape

\begin{table}
\begin{footnotesize}
\caption{Cpu times (t) from frequency allocation problems. Best cpu time is in bold.
The s and g prefixes stand for scen and graph respectively.} 

\setlength{\tabcolsep}{4pt}
\begin{tabular}{|c|c|c|c|c|c|c|c|c|c|c|c|c|c|c|c|c|}
\hline
 &  & &$d/wdeg$&$d/wdeg$&$d/wdeg$& &$Node$&$Impact$& & $alldel$ & $alldel$ & $alldel$& &$fully$&$fully$ &$fully$\\
Instance & & $d/wdeg$ & $r.probe$  & $RSC$ & $r.probe$ & $Impact$ & $Impact$ & $RSC$& $alldel$ & $r.probe$ & $RSC$ & $r.probe$&$fully$ & $r.probe$ & $RSC$ & $r.probe$\\
  & & & & & $RSC$ &  &  & & & & & $RSC$& & & & $RSC$\\
\hline 
s2-f25 & t & {\bf 7,4} & 29,2 & 14,2 & 23,6 & 14,2 & 19,5 & 19,5 & 9,3 & 28,5 & 9,8 & 22,7 & 11,1 & 29,8 & 9,1 & 29,4\\
(unsat) & n & 1195 & 16317 & 2651 & 14548 & 2088 & 2091 & 2091 & 1689 & 16231 & 1579 & 14321 & 1744 & 16672 & 1388 & 16211\\ \hline
s3-f10 & t & 2,2 & 36,5 & 10,2 & 33,7 & $>$ 1h & $>$ 1h & $>$ 1h & 2,3 & 38,7 & 5,5 & 30 & {\bf 1,2} & 37,2 & 10,4 & 37,2\\
(sat) & n & 724 & 18119 & 728 & 17781 & -- & -- & -- & 900 & 18312 & 941 & 17147 & 472 & 927 & 631 & 18891\\ \hline
s3-f11 & t & 9,6 & 41,4 & 9,9 & 36,2 & $>$1h & $>$1h & $>$1h & {\bf 5,3} & 47,5 & 10,4 & 39,9 & 9,6 & 47,7 & 11,7 & 48,7\\
(unsat) & n & 1078 & 18728 & 861 & 17211 & -- & -- & -- & 641 & 18862 & 889 & 17865 & 1078 & 18993 & 1546 & 19944\\ \hline
g8-f10 & t & 15 & 72,5 & 45,2 & 62,4 & $>$1h & $>$1h & $>$1h & 21,3 & 76,5 & 14,4 & 60,7 & {\bf 10,9} & 77,4 & 15,8 & 66,4\\
(sat) & n & 4193 & 26535 & 6018 & 22739 & -- & -- & -- & 6877 & 27781 & 3887 & 23005 & 3428 & 27193 & 4127 & 24162\\\hline
g8-f11 & t & 7 & 62,7 & 10,5 & 54,2 & $>$ 1h & $>$ 1h & $>$ 1h & 1,6 & 60,3 & 1,7 & 55,3 & {\bf 0,8} & 61 & 1,1 & 58,3\\
(unsat) & n & 1450 & 24244 & 940 & 23348 & -- & -- & -- & 224 & 23878 & 455 & 23668 & 107 & 23979 & 105 & 24138\\\hline
g14-f27 & t & {\bf 18,8} & 48,4 & 82,5 & 49,1 & 53,9 & 216,7 & 217,1 & 28,8 & 48,3 & 70,1 & 60,2 & 39,8 & 52,5 & 89,3 & 52,7\\
(sat) & n & 12251 & 28337 & 13106 & 27785 & 6284 & 6284 & 6284 & 18143 & 28019 & 40211 & 38901 & 20820 & 29211 & 47655 & 31925\\\hline
g14-f28 & t & 75,3 & 43,3 & 18,2 & 37,2 & $>$ 1h & $>$ 1h & $>$ 1h & {\bf 0,4} & 60,5 & 57,3 & 55,7 & 46,4 & 51,5 & 57,1 & 53,1\\
(unsat) & n & 33556 & 22303 & 1459 & 19544 & -- & -- & -- & 99 & 29928 & 30239 & 29327 & 16397 & 20389 & 24356 & 28874\\ \hline
s11 & t & 5,5 & 118,1 & 141,2 & 157,5 & 29,3 & 210,6 & 224,8 & {\bf 4} & 120,5 & 56,2 & 97,4 & 4,3 & 127,8 & 120,3 & 181,3\\
(sat) & n & 1024 & 35097 & 959 & 29391 & 834 & 833 & 833 & 947 & 35788 & 1540 & 29080 & 853 & 36611 & 780 & 35610\\ \hline
s11-f12 & t & 6,6 & 56,2 & 4,8 & 51,5 & 22,4 & 25,7 & 25,8 & 3,7 & 54,1 & 3,9 & 48,3 & {\bf 3,1} & 54,3 & 3,5 & 52,6\\
(unsat) & n & 1102 & 24158 & 981 & 22893 & 421 & 421 & 421 & 566 & 23661 & 989 & 21775 & 386 & 23865 & 977 & 22798\\\hline
s11-f11 & t & 6,8 & 55,6 & 4,7 & 51,1 & 22,1 & 25,7 & 25,8 & 3,7 & 53,4 & 3,8 & 48,5 & {\bf 3,1} & 53,7 & 3,6 & 51,8\\
(unsat) & n & 1102 & 23555 & 981 & 22751 & 421 & 421 & 421 & 522 & 23101 & 989 & 21557 & 386 & 23566 & 977 & 22564\\ \hline
s11-f10 & t & 3,5 & 56,7 & 4,8 & 52,4 & $>$ 1h & $>$ 1h & $>$ 1h & 4,5 & 58,2 & {\bf 3,3} & 55,8 & 4,5 & 59,7 & 4,6 & 59,4\\
(unsat) & n & 490 & 23131 & 498 & 22891 & -- & -- & -- & 556 & 23664 & 376 & 24077 & 528 & 23512 & 631 & 24972\\ \hline
s11-f9 & t & 14,3 & 71,7 & 18,1 & 65,6 & $>$ 1h & $>$ 1h & $>$ 1h & 16,4 & 72,4 & 18,9 & 65,2 & {\bf 12,1} & 72,8 & 17,1 & 71,1\\
(unsat) & n & 1412 & 24261 & 1384 & 23441 & -- & -- & -- & 1906 & 24547 & 1753 & 23287 & 1156 & 24781 & 1150 & 24763\\\hline
s11-f8 & t & {\bf 21,2} & 87,1 & 44,7 & 79,6 & $>$ 1h & $>$ 1h & $>$ 1h & 26,3 & 89,1 & 40,1 & 76,8 & 26,1 & 89,5 & 45 & 83,5\\
(unsat) & n & 2112 & 28083 & 2897 & 24892 & -- & -- & -- & 2526 & 27867 & 3192 & 24682 & 2181 & 27944 & 2784 & 27443\\\hline
s11-f7 & t & 133,7 & 189,9 & 211,2 & 201,6 & $>$ 1h & $>$ 1h & $>$ 1h & {\bf 130,6} & 191,4 & 166,8 & 221,8 & 137,7 & 198,5 & 160,2 & 203,5\\
(unsat) & n & 12777 & 39469 & 20154 & 42345 & -- & -- & -- & 13205 & 39557 & 14886 & 45388 & 12777 & 24689 & 15017 & 42167\\\hline
s11-f6 & t & 391,4 & 402,9 & 412,7 & 416,8 & $>$ 1h & $>$ 1h & $>$ 1h & {\bf 307,6} & 488,4 & 465,2 & 479,8 & 330,4 & 330,4 & 301,4 & 320,6\\
(unsat) & n & 34714 & 61523 & 40892 & 62557 & -- & -- & -- & 27949 & 63447 & 37954 & 69432 & 28947 & 29930 & 19236 & 29084\\ \hline\hline
Averaged cpu time & t & 47,9 & 91,5 & 68,8 & 91,5 & -- & -- & -- & {\bf 37,7} & 99,1 & 61,8 & 94,5 & 42,7 & 89,5 & 56,8 & 91,3\\ \hline

\end{tabular}

\label{table:rlfap} 
\end{footnotesize}
\end{table}

\endlandscape

\newpage

Although the Impact heuristic seems to make a better exploration
of the search tree on some easy instances (like s2-f25, g14-f27,
s11, s11-f12), it is clearly slower compared to
conflict-driven heuristics. This is mainly because the process of
impact initialization is time consuming. On hard instances, the
Impact heuristic has worse performance and in some cases it cannot
solve the problem within the time limit on all instances. In general we observed
that impact based heuristics cannot handle efficiently problems
which include variables with relatively large domains. Some RLFA
problems, for example, have 680 variables with up to 44 values
in their domains.

Node Impact and its variation, ``Impact RSC'', are strongly
related, and this similarity is depicted in the results. As
mentioned in Section~\ref{sec:heuristics}, Node Impact computes
the accurate impacts and the ``RSC'' heuristic computes the
reduction in the search space, after the application of Restricted
Singleton Consistency. Since node impact computation also uses
Restricted Singleton Consistency (it subsumes it), these
heuristics differ only in the measurement function that assigns
impacts to variables. Hence, when they are used to break ties on
the Impact heuristic, they usually make similar decisions.

When ``RSC'' is used as a tie breaker for conflict-driven
heuristics, results show that it does not offer significant
changes in the performance. So we have excluded it from the
experiments that follow in the next sections, except for the
dom/wdeg + RSC combination.

Concerning ``random probing'', although experiments in
\cite{grim07} show that it has often better performance when
compared to simple \emph{dom/wdeg}, our results show that this is
not the case when \emph{dom/wdeg} is combined with a geometric
restart strategy. Even on hard instances, where the computation
cost of random probes is small compared to the total search
cost, results show that \emph{dom/wdeg} and its variations are
dominant. Moreover, the combination of ``random probing'' with any
other conflict-driven variation heuristic (``alldel'' or ``fully
assigned'') does not result in significant changes in the
performance. Thus, for the next experiments we have kept only the
``random probing'' and \emph{dom/wdeg} combination.

Finally, among the three conflict-driven variations, \emph{``alldel''} seems to display slightly
better performance on this set of instances.

\subsection{Structured  and patterned instances}

\label{subsec:struct}

This set of experiments contains instances from academic problems
(langford), some real world instances from the ``driver'' problem
and 6 patterned instances from the graph coloring problem. The constraint graphs of the latter are randomly generated but the structure of the constraints follows a specific pattern as they are all binary inequalities. Since some of the variations presented in the previous paragraph
(Table~\ref{table:rlfap}) were shown to be less interesting, we
have omitted their results from the next tables.

Results in Table~\ref{table:struct} show that the behavior of the
selected heuristics is close to the behavior that we observed in
RLFA problems.  Conflict-driven variations are again dominant
here. The \emph{dom/wdeg} heuristic has in most cases the best
performance, followed by ``alldel'' and ``fully assigned''. Impact based heuristics have by far the
worst performance. Random probing again seems to be an overhead as it increases both run times and nodes visits.

\begin{table}[t]
\begin{footnotesize}
\caption{Cpu times (t), and nodes (n) from structured and patterned problems. Best cpu time is in bold.} 

\setlength{\tabcolsep}{4pt}
\begin{tabular}{|c|c|c|c|c|c|c|c|c|c|}
\hline
Instance &  & $d/wdeg$ &$d/wdeg$&$d/wdeg$& $Impact$ & $Node$ &$Impact$& $alldel$ & $fully$\\
& & & $r.probe$  & $RSC$ & & $Impact$ & $RSC$& $by$ $\#del$ & $assigned$\\
\hline 
langford- & t & {\bf 42,8} & 48,5 (44,1) & 52,2 & 65,5 & 70 & 73,8 & 46,9 & 48,2 \\
2-9(unsat) & n & 65098 & 64571 (59038) & 68901 & 73477 & 52174 & 53201 & 62171 & 60780 \\ \hline
langford- & t & {\bf 364,5} & 380 (374,2) & 431,2 & 406,9 & 660,6 & 530,7 & 402,2 & 395,2 \\
2-10(unsat) & n & 453103 & 422742 (417227) & 481909 & 458285 & 494407 & 479092 & 435599 & 428681 \\ \hline
langford- & t & {\bf 584,8} & 673,2 (621) &  632,8 & 1094 & 1917 & 1531 & 726,6 & 676,8 \\
3-11(unsat) & n & 140168 & 134133 (126991) & 140391 & 174418 & 200558 & 187091 & 141734 & 138919 \\ \hline
langford- & t & {\bf 65,9} & 238,2 (65,3) & 101,3 & 183,4 & 289,3 & 301,1 & 106,7 & 70,3 \\
4-10(unsat) & n & 5438 & 14024 (4582) & 5099 & 9257 & 9910 & 9910 & 7362 & 5031 \\ \hline
driver-& t & 13,6 & 43,1 (0,7) & 31,2 & 27,8 & 31,2 & 31,1 & {\bf 1,3} & 1,4 \\
8c (sat) & n & 4500 & 9460 (420) & 3110 & 431 & 429 & 429 & 660 & 632 \\ \hline
driver-9& t & 262,3 & 305,2 (219,7) & 201,1 & $>1h$ & 1409 & 2121 & {\bf 123,5} & 167,9 \\
(sat) & n & 58759 & 58060 (46413) & 18581 & -- & 19668 & 60291 & 13657 & 20554 \\ \hline
will199-5& t & {\bf 1,4} & 17 (1,7) & 5,2 & $>1h$ & $>1h$ & $>1h$ & 1,7 & 2,1 \\
(unsat) & n & 577 & 13060 (726) & 650 & -- & -- & -- & 538 & 582 \\ \hline
will199-6& t & 15,8 & 42,9 (21,9) & 30,1 & $>1h$ & $>1h$ & $>1h$ & {\bf 12,7} & 13,4 \\
(unsat) & n & 4288 & 22792 (5763) & 4582 & -- & -- & -- & 2852 & 2846 \\ \hline
ash608-4& t & 3,3 & 20,1 (1,8) & 81,3 & 35,1 & 136,2 & 123,3 & 2,6 & {\bf 1,2} \\
(sat) & n & 3146 & 21346 (1823) & 2291 & 3860 & 2452 & 2293 & 2586 & 1266 \\ \hline
ash958-4& t & 12,8 & 36,8 (3) & 299,2 & 111,4 & $>1h$ & $>1h$ & 11,6 & {\bf 1,2} \\
(sat) & n & 8369 & 27322 (1992) & 3870 & 5105 & -- & -- & 7399 & 1266 \\ \hline
ash313-5& t & {\bf 18,2} & 134,7 (18,4) & 43,2 & 172,2 & 442,1 & 489,7 & 19,4 & 19 \\
(unsat) & n & 512 & 10204 (512) & 512 & 512 & 512 & 512 & 512 & 512 \\ \hline
ash313-7& t & {\bf 828,4} & 1011 (809,6) & 1271 & 1015 & $>1h$ & $>1h$ & 995,7 & 1056 \\
(unsat) & n & 20587 & 35135 (19990) & 20139 & 20539 & -- & -- & 20411 & 20406 \\ \hline\hline
Averaged cpu time & t & {\bf 184,4} & 245,8 & 264,9 & -- & -- & -- & 204,2 & 204,3 \\\hline
\end{tabular}

\label{table:struct} 
\end{footnotesize}
\end{table}

\begin{table}
\begin{footnotesize}
\caption{Cpu times (t), and nodes (n) from random problems. Best cpu time is in bold.} 

\setlength{\tabcolsep}{4pt}
\begin{tabular}{|c|c|c|c|c|c|c|c|c|c|}
\hline
Instance &  & $d/wdeg$ &$d/wdeg$&$d/wdeg$& $Impact$ & $Node$ &$Impact$& $alldel$ & $fully$\\
& & & $r.probe$  & $RSC$ & & $Impact$ & $RSC$& $by$ $\#del$ & $assigned$\\
\hline 
ehi-85-0 & t & 2,1 & 94,2 (0,15) & 2,7 & 11,7 & 12,1 & 12 & {\bf 0,15} & 1,2\\
(unsat) & n & 722 & 8005 (4) & 61 & 3 & 3 & 3 & 4 & 149 \\ \hline
ehi-85-2 & t & {\bf 1} & 101,6 (0,15) & 2,4 & 11,8 & 12,4 & 12,4 & 5,6 & 1,1 \\
(unsat) & n & 248 & 7944 (5) & 12 & 4 & 4 & 4 & 650 & 145 \\ \hline
geo50-d4- & t & 334,9 & 526 (490,6) & 311,3 & $>1h$ & $>1h$ & $>1h$ & 280 & {\bf 129,2} \\
75-2(sat) & n & 50483 & 88615 (76247) & 46772 & -- & -- & -- & 42946 & 18545  \\ \hline
frb30-15-1 & t & {\bf 10,5} & 42 (15,4) & 13,2 & 66,4 & 295,6 & 375,6 & 20,5 & 15,6 \\
(sat) & n & 3557 & 15833 (4426) & 3275 & 17866 & 71052 & 85017 & 6044 & 4493 \\ \hline
frb30-15-2 & t & 63,7 & 123,6 (97,8) & 55,4 & 273,4 & {\bf 5,4} & 391,3 & 86,8 & 91,2 \\
(sat) & n & 21330 & 38765 (27458) & 20019 & 79936 & 1306 & 81911 & 26596 & 26296 \\ \hline
40-8-753- & t & 76,5 & 70,9 (45,9) & 60,4 & 2117 & 404,5 & 931,3 & {\bf 50,5} & 486,3 \\
0,1 (sat) & n & 21164 & 21369 (13422) & 15239 & 523831 & 67979 & 180281 & 13823 & 127686 \\ \hline
40-11-414- & t & 1192 & 1261 (1234) & 1219 & $>1h$ & $>1h$ & $>1h$ & 1178 & {\bf 1162} \\
0,2 (unsat) & n & 336691 & 354778 (345212) & 345886 & -- & -- & -- & 346368 & 332844 \\ \hline
40-16-250- & t & 2919 & 2928 (2895) & 3172 & $>1h$ & $>1h$ & $>1h$ & {\bf 2893} & 3038 \\
0,35 (unsat) & n & 741883 & 755386 (743183) & 750910 & -- & -- & -- & 747757 & 764989 \\ \hline
40-25-180- & t & 2481 & 2689 (2632) & 2878 & $>1h$ & $>1h$ & $>1h$ & {\bf 2340} & 2606 \\
0,5 (unsat) & n & 373742 & 402266 (385072) & 390292 & -- & -- & -- & 349685 & 389603 \\ \hline\hline
Averaged cpu time & t & 786,7 & 870,7 & 857,1 & -- & -- & -- & {\bf 761,6} & 836,7 \\\hline
\end{tabular}

\label{table:random1} 
\end{footnotesize}
\end{table}

\begin{table}
\begin{footnotesize}
\caption{Cpu times (t), and nodes (n) from problems with non-binary constraints. Best cpu time is in bold.} 

\setlength{\tabcolsep}{4pt}
\begin{tabular}{|c|c|c|c|c|c|c|c|c|c|c|}
\hline
Instance &  & $d/wdeg$ &$d/wdeg$&$d/wdeg$& $Impact$ & $Node$ &$Impact$& $alldel$ & $fully$\\
& & & $r.probe$  & $RSC$ & & $Impact$ & $RSC$& $by$ $\#del$ & $assigned$\\
\hline 
cc-10-10-2 & t & 31 & 40,6 (30,3) & 47,7 & 31,3 & 193,1 & 219,2 & {\bf 29,9} & 33,1 \\
(unsat) & n & 16790 & 20626 (15800) & 16544 & 16161 & 10370 & 10233 & 15639 & 15930 \\ \hline
cc-12-12-2 & t & 50,7 & 67,6 (14,3) & 79,3 & 65 & 523,6 & 555,6 & {\bf 49,1} & 54,3 \\
(unsat) & n & 16897 & 19429 (49780) & 16596 & 21532 & 13935 & 13564 & 16292 & 16135 \\ \hline
cc-15-15-2 & t & 98,6 & 125 (94,5) & 159,7 & {\bf 91,3} & 1037 & 1134 & 103,6 & 102,1 \\
(unsat) & n & 16948 & 20166 (14881) & 16674 & 16437 & 10374 & 10012 & 15741 & 15945 \\ \hline
series-16 & t & {\bf 147,3} & 543,9 (516,5) & 177,6 & $>1h$ & $>1h$ & $>1h$ & $>1h$ & $>1h$ \\
(sat) & n & 49857 & 155102 (146942) & 51767 & -- & -- & -- & -- & -- \\ \hline
series-18 & t & $>1h$ & $>1h$ & $>1h$ & $>1h$ & $>1h$ & $>1h$ & $>1h$ & $>1h$ \\
(sat) & n & -- & -- & -- & -- & -- & -- & -- & -- \\ \hline
renault-mod-0 & t & 1285 & $>1h$ & 2675 & $>1h$ & $>1h$ & $>1h$ & 1008 & {\bf 776,2} \\
(sat) & n & 288 & -- & 251 & -- & -- & -- & 166 & 179 \\ \hline
renault-mod-1 & t & 2126 & $>1h$ & 2283 & $>1h$ & $>1h$ & $>1h$ & {\bf 431,4} & 785,4 \\
(unsat) & n & 474 & -- & 469 & -- & -- & -- & 161 & 234 \\ \hline
renault-mod-3 & t & 2598 & $>1h$ & 2977 & $>1h$ & $>1h$ & $>1h$ & 993,5 & {\bf 435,7} \\
(unsat) & n & 546 & - & 475 & -- & -- & -- & 203 & 176 \\ \hline
\end{tabular}

\label{table:nonbinary} 
\end{footnotesize}
\end{table}

\begin{table}
\begin{footnotesize}
\caption{Cpu times (t), and nodes (n) from boolean problems. Best cpu time is in bold.} 

\setlength{\tabcolsep}{4pt}
\begin{tabular}{|c|c|c|c|c|c|c|c|c|c|c|}
\hline
Instance &  & $d/wdeg$ &$d/wdeg$&$d/wdeg$& $Impact$ & $Node$ &$Impact$& $alldel$ & $fully$\\
& & & $r.probe$  & $RSC$ & & $Impact$ & $RSC$& $by$ $\#del$ & $assigned$\\
\hline 
jnh01 & t & 10,2 & 95,2 (3,5) & 14,2 & {\bf 2,2} & 13,2 & 13 & 4,2 & 6,2 \\
(sat) & n & 970 & 5215 (362) & 515 & 100 & 100 & 100 & 481 & 692 \\ \hline
jnh17 & t & 3,1 & 57,3 (0,5) & 18,8 & 1,9 & 10,1 & 9,9 & {\bf 1,4} &  1,5 \\
(sat) & n & 477 & 4914 (189) & 1233 & 132 & 131 & 131 & 216 & 204 \\ \hline
jnh201 & t & 3 & 79,6 (2,1) & 3,3 & 2,6 & 11 & 10,7 & {\bf 1,13} & 1,14 \\
(sat) & n & 336 & 5222 (121) & 168 & 177 & 180 & 180 & 179 & 178 \\ \hline
jnh301 & t & 33,4 & 121 (14,5) & 38,2 & {\bf 2,2 } & 5,7 & 5,5 & 7 & 8 \\
(sat) & n & 2671 & 6144 (1488) & 1541 & 110 & 108 & 108 & 608 & 787 \\ \hline
aim-50-1- & t & 0,15 & 0,43 (0,13) & 0,21 & 0,82 & 0,49 & 0,5 & {\bf 0,07} & 0,08 \\
6-unsat-2 & n & 1577 & 6314 (1404) & 1412 & 6774 & 474 & 474 & 691 & 474 \\ \hline
aim-100-1- & t & 0,34 & 1,47 (1,05) & 1,42 & 91 & 4,3 & 6,3 & {\bf 0,16} & 0,2 \\
6-unsat-1 & n & 3592 & 17238 (10681) & 7932 & 697503 & 2338 & 3890 & 1609 & 1229  \\ \hline
aim-200-1- & t & 0,76 & 1,28 (0,47) & 2,1 & 1,66 & 1,9 & 1,8 & {\bf 0,24} & 0,26 \\
6-sat-1 & n & 4665 & 11714 (3236) & 1371 & 4747 & 213 & 213 & 1756 & 1442 \\ \hline
aim-200-1- & t & 1,9 & 3,2 (2,4) & 5,3 & 105,9 & 4,8 & 8,5 & {\bf 0,19} & 0,23 \\
6-unsat-1 & n & 12748 & 26454 (16159) & 28548 & 436746 & 1615 & 3654 & 1255 & 1093 \\ \hline
pret-60- & t & 1255 & 1385 (1385) & $>1h$ & 3589 & $>1h$ & $>1h$ & {\bf 1027} & 1108 \\
25 (unsat) & n & 44,6M & 44,777M (43,773M) & -- & 95,4M & -- & -- & 42,5M & 43,8M \\ \hline
dubois-20 & t & 1196 & 1196 (1196) & $>1h$ & $>1h$ & $>1h$ & $>1h$ & {\bf 1004} & 1245 \\
(unsat) & n & 44,9M & 44,461M (44,457M) & -- & -- & -- & -- & 40,5M & 43,8M \\ \hline\hline
Averaged cpu time & t & 250,3 & 294 & 857,1 & -- & -- & -- & {\bf 204,5} & 237 \\\hline
\end{tabular}

\label{table:boolean} 
\end{footnotesize}
\end{table}

\subsection{Random instances}

\label{subsec:rand}

In this set of experiments we have selected some quasi-random
instances which contain some structure (``ehi'' and ``geo''
problems) and also some purely random instances, generated
following Model RB and Model D.

Model RB instances (frb30-15-1 and frb30-15-2) are random
instances forced to be satisfiable. Model D instances are
described by four numbers $<$\emph{n,d,e,t}$>$. The first number
$n$ corresponds to the number of variables. The second number $d$
is the domain size and $e$ is the number of constraints. $t$ is
the tightness, which denotes the probability that a pair of values
is allowed by a relation.

Results are presented in Table~\ref{table:random1}. All the
conflict-driven heuristics (\emph{dom /wdeg}, ``alldel'' and
``fully assigned'') have much better
cpu times compared to impact based heuristics. In pure random
problems the ``alldel'' heuristic has the best cpu times, while in
quasi-random instances the three conflict-driven heuristics share
a win. Random probing can slightly improve the performance of \emph{dom/wdeg} on Model D problems but it is an overhead on the rest of the instances.

\subsection{Non-binary instances}

\label{subsec:nonbinary}

In this set of experiments we have included problems with
non-binary constraints. The first three instances are from the
chessboard coloration problem. This problem is the task of
coloring all squares of a chessboard composed by $r$ rows and $c$
columns. There are exactly $n$ available colors and the four
corners of any rectangle extracted from the chessboard must not be
assigned the same color. Each instance is denoted by
cc-$r$-$c$-$n$. These instances have maximum arity of 4.

The next two instances are from the academic problem ``All
Interval Series'' (See prob007 at http://www.csplib.org) which
have maximum arity of 3, while the last three instances are from a
Renault Megane configuration problem where symbolic domains have
been converted to numeric ones. The renault instances have maximum
arity of 10.

Results are presented in Table~\ref{table:nonbinary}. Here again
the conflict-driven heuristics have the best performance in most
cases. The Impact based heuristics have the best cpu performance
in two instances (cc-15-15-2 and series-16), but on the other hand
they cannot solve 4 instances within the time limit.

We must also note here that although the ``node impact'' and
``impact RSC'' heuristics are slow on chessboard coloration
instances, they visit less nodes. In general, with impact based
heuristics there are cases where we can have a consistent
reduction in number of visited nodes, albeit at the price of
increasing the running time.

Random probing is very expensive for non-binary problems, especially when the arity of the constraints is large and the cost of constraint propagation is high. As a result, adding random probing forced the solver to time out on many instances.

\subsection{Boolean instances}

\label{subsec:bool}

This set of experiments contains instances involving only Boolean
variables and non-binary constraints. We have selected a
representative subset from Dimacs problems. To be precise, we have
included a subset of the ``jnhSat'' collection which includes the
hardest instances from this collection, 4 randomly selected instances from the
``aim'' set, where all problems are relatively easy to solve,
and the first instance from the
``pret'' and ``dubois'' sets, which include very hard instances.
All the selected instances have constraint arity of 3, except for
the ``jnhSat'' instances which have maximum arity of 14.

Results from these experiments can be found in
Table~\ref{table:boolean}. The behavior of the evaluated
heuristics in this data set is slightly different from the
behavior that we observed in previous problems. Although
conflict-driven heuristics again display the best overall
performance, impact based heuristics are in some cases faster.

The main bottleneck that impact based heuristics have, is
the time consuming initialization process. On Boolean instances,
where the variables have binary domains, the cost for the
initialization of impacts is small. And this can significantly
increase the performance of these heuristics.


Among the conflict-driven heuristics, the ``alldel'' heuristic is
always better than its competitors. We recall here that in this
heuristic constraint weights are increased by the size of the
domain reduction. Hence, on binary instances constraint weights
can be increased at minimum by one and at maximum by two (in each
DWO).

The same good performance of the ``alldel'' heuristic was also
observed in 30 additional instances from the Dimacs problem class (``aim'' instances)
not shown here. These extended experiments showed that this way of
incrementing weights seems to work better on Boolean problems
where the deletion of a single value is of greater importance
compared to problems with large domains, i.e. it is more likely to
lead to a DWO.

\subsection{The effect of restarts on the results}

In all the experiments reported in the previous sections we
followed a geometric restart policy. This policy were introduced in \cite{walsh99ijcai}
and it has been shown to be very effective.
However, different restart
policies can be applied within the search algorithm, or we can
even discard restarts in favor of a single search run. In order to
check how the selected restart policy affects the performance of
the evaluated variable ordering heuristics, we ran some additional
experiments.

Apart from the geometric restart policy which we used on the
previous experiments, we also tried an arithmetic restart policy.
In this policy the initial number of allowed backtracks for the
first run has been set to 10 and at each new run the number of
allowed backtracks increases also by 10. We have also tested the
behavior of the heuristics without the use of any restarts.

Selected results are depicted in Table~\ref{table:restarts}. Unsurprisingly,
results show that the arithmetic restart policy is clearly
inefficient. On instances that can be solved within a small number
of restarts (like \emph{scen11, ehi-85-297-0, rb30-15-1} and
\emph{ash958GPIA-4}), the differences between the arithmetic and
the geometric restart policies are small. But, when some problem
(like \emph{scen11-f7, aim-200-1-6, langford-4-10} and
\emph{cc-12-12-2}) requires a large number of restarts to be
solved, the geometric restart policy clearly outperforms the
arithmetic one. Importantly for the purposes of this paper, this behavior is independent of the selected
variable ordering heuristic.

Comparing search without restart to the geometric restart policy,
we can see that the latter is more efficient some instances.
But in general restarts are necessary to solve
very hard problems. Importantly, the relative behavior of the
conflict-driven heuristics compared to impact based heuristics is
not significantly affected by the presence or absence of restarts.
That is, the conflict-driven heuristics are always faster than the
impact based ones, with or without restarts. Some small
differences in the relative performance of the conflict-driven
heuristics can be noticed when no restarts are used, but they
generally have similar cpu times. \emph{Random
probing} seems to work better with no restarts, in accordance with the results and conjectures in \cite{grim08}, but this
small improvement is not enough for it to become more efficient
than the \emph{dom/wdeg}, ``\emph{alldel}'' and ``\emph{fully
assigned}'' heuristics.

\label{subsec:restarts}

\begin{table}
\begin{footnotesize}
\caption{Cpu times for the three selected restart policies: without restarts, arithmetic restarts and geometric restarts. Best cpu time is in bold.} 

\setlength{\tabcolsep}{4pt}
\begin{tabular}{|c|c|c|c|c|c|c|c|c|c|c|}
\hline
Instance & restart & $d/wdeg$ &$d/wdeg$&$d/wdeg$& $Impact$ & $Node$ &$Impact$& $alldel$ & $fully$\\
& policy & & $r.probe$  & $RSC$ & & $Impact$ & $RSC$& $by$ $\#del$ & $assigned$\\
\hline 
scen11 & no restart & 42,5 & 102,2 & 148,3 & $>1h$ & $>1h$ & $>1h$ & 112,4 & {\bf 41,3} \\
(sat) & arithmetic & 8 & 109,5 & 142,7 & 29 & 211,3 & 218,3 & {\bf 4} & 4,5 \\
 & geometric & 5,5 & 118,1 & 141,2 & 29,3 & 210,6 & 224,8 & {\bf 4} & 4,3 \\ \hline
scen11-f7 & no restart & $>1h$ & {\bf 109} & $>1h$ & $>1h$ & $>1h$ & $>1h$ & $>1h$ & $>1h$ \\
(unsat) & arithmetic & 1848 & {\bf 1464} & 1991 & $>1h$ & $>1h$ & $>1h$ & 3207 & 2164 \\
 & geometric & 133,7 & 189,9 & 211,2 & $>1h$ & $>1h$ & $>1h$ & {\bf 130,6} & 137,7 \\ \hline
aim-200-1-6 & no restart & 4,8 & 1,5 & 4,7 & 2,3 & 2,8 & 3,1 & {\bf 0,28} & 0,31 \\
(unsat) & arithmetic & 81,4 & 150,3 & 212,7 & 124,8 & 9,4 & 9,2 & 0,39 & {\bf 0,27} \\
 & geometric & 1,9 & 3,2 & 5,3 & 105,9 & 4,8 & 8,5 & {\bf 0,19} & 0,23 \\ \hline
 ehi-85-297-0 & no restart & 17,1 & 90,4 & 7,1 & 11,8 & 12,2 & 12,4 & {\bf 0,16} & 1,9 \\
(unsat) & arithmetic & 2 & 102,2 & 2,8 & 12,8 & 12,4 & 12,3 & {\bf 0,15} & 1,18 \\
 & geometric & 2,1 & 94,2 & 2,7 & 11,7 & 12,1 & 12 & {\bf 0,15} & 1,2 \\ \hline
 frb30-15-1 & no restart & {\bf 3,2} & 30,1 & 3,6 & 152,6 & 215,1 & 201,5 & 7,1 & 3,8 \\
(sat) & arithmetic & 15,9 & 149,2 & {\bf 15,1} & 303,9 & 626,1 & 532,5 & 184,3 & 201,2 \\
 & geometric & {\bf 10,5} & 42 & 13,2 & 66,4 & 295,6 & 375,6 & 20,5 & 15,6 \\ \hline
  langford-4-10 & no restart & {\bf 16,2} & 193,7 & 24,6 & 59,1 & 74,6 & 79,3 & 24,1 & 20,8 \\
(unsat) & arithmetic & {\bf 521,1} & 749,7 & 557,2 & 2579 & 904,1 & 1293 & 1011 & 744,9 \\
 & geometric & {\bf 65,9} & 238,2 & 101,2 & 183,4 & 289,3 & 301,1 & 106,7 & 70,3 \\ \hline
  cc-12-12-2 & no restart & 17 & 27,6 & 25,4 & {\bf 16,4} & 115,9 & 98,2 & 17,9 & 17,2 \\
(unsat) & arithmetic & 2939 & {\bf 1976} & $>1h$ & $>1h$ & $>1h$ & $>1h$ & 2501 & 2589 \\
 & geometric & 50,7 & 67,6 & 79,3 & 65 & 523,6 & 555,6 & {\bf 49,1} & 54,3 \\ \hline
   ash958GPIA-4 & no restart & 10,4 & 35,6 & 162,2 & 383,7 & $>1h$ & $>1h$ & {\bf 6,3} & 6,4 \\
(sat) & arithmetic & 13 & 36,2 & 310,2 & 118,3 & $>1h$ & $>1h$ & 14 & {\bf 10,7} \\
 & geometric & 12,8 & 36,8 & 299,2 & 111,4 & $>1h$ & $>1h$ & 11,6 & {\bf 1,2} \\ \hline
\end{tabular}

\label{table:restarts} 
\end{footnotesize}
\end{table}

\subsection{Using random value ordering}

\label{subsec:valueOrdering}

As noted at the beginning of Section~\ref{sec:experiments},
all the experiments were ran with a lexicographic value ordering.
In order to check if this affects the performance of
the evaluated variable ordering heuristics, we have ran some additional experiments.
In these experiments we study the performance of the heuristics when  random value ordering is used.

Selected results are depicted in Table~\ref{table:valueOrdering} where
we show cpu times for both random and lexicographic value ordering. Concerning
the random value ordering, all the results presented here are averaged values for 50 runs.
Looking at the results and comparing the performance of the heuristics under the different value orderings, we can see some differences in cpu time. However, the relative behavior of the
conflict-driven heuristics compared to impact based heuristics is
not significantly affected by the use of lexicographic or random value ordering.

\begin{table}
\begin{footnotesize}
\caption{Cpu times for the two different value orderings: lexicographic and random.
 Best cpu time for each ordering is in bold.} 

\setlength{\tabcolsep}{4pt}
\begin{tabular}{|c|c|c|c|c|c|c|c|c|c|c|}
\hline
Instance & value & $d/wdeg$ &$d/wdeg$&$d/wdeg$& $Impact$ & $Node$ &$Impact$& $alldel$ & $fully$\\
& ordering & & $r.probe$  & $RSC$ & & $Impact$ & $RSC$& $by$ $\#del$ & $assigned$\\
\hline 
scen11-f7 & random & 161 & 232,5 & 191,3 & $>1h$ & $>1h$ & $>1h$ & {\bf 157,2} & 178,9 \\
(unsat) & lexico & 133,7 & 189,9 & 211,2 & $>1h$ & $>1h$ & $>1h$ & {\bf 130,6} & 137,7 \\ \hline
aim-200-1-6 & random & 2,3 & 2,3 & 6,2 & 11,9 & 6,4 & 6,1 & {\bf 0,18} & 0,23 \\
(unsat) & lexico & 1,9 & 3,2 & 5,3 & 105,9 & 4,8 & 8,5 & {\bf 0,19} & 0,23 \\ \hline
 ehi-85-297-0 & random & 1,3 & 3,5 & 5,1 & 11,4 & 11,8 & 11,9 & {\bf 0,16} & 0,8 \\
(unsat) & lexico & 2,1 & 94,2 & 2,7 & 11,7 & 12,1 & 12 & {\bf 0,15} & 1,2 \\ \hline
 frb30-15-1 & random & 39,7 & 52,8 & 27,5 & 120,9 & 132,4 & 123,6 & 32,3 & {\bf 28,2} \\
(sat) & lexico & {\bf 10,5} & 42 & 13,2 & 66,4 & 295,6 & 375,6 & 20,5 & 15,6 \\ \hline
  langford-4-10 & random & {\bf 61,2} & 229,8 & 75,4 & 155,7 & 255,7 & 249,6 & 280,5 & 83,8 \\
(unsat) & lexico & {\bf 65,9} & 238,2 & 101,2 & 183,4 & 289,3 & 301,1 & 106,7 & 70,3 \\ \hline
  cc-12-12-2 & random & 55,6 & 74,5 & 82,4 & {\bf 51,2} & 423,9 & 437,2 & 55,8 & 54,8 \\
(unsat) & lexico & 50,7 & 67,6 & 79,3 & 65 & 523,6 & 555,6 & {\bf 49,1} & 54,3 \\ \hline
   ash958GPIA-4 & random & 5,3 & 35,6 & 242,2 & 106,6 & 515,4 & 450,1 & {\bf 3,8} & 3,9 \\
(sat) & lexico & 12,8 & 36,8 & 299,2 & 111,4 & $>1h$ & $>1h$ & 11,6 & {\bf 1,2} \\ \hline
\end{tabular}

\label{table:valueOrdering} 
\end{footnotesize}
\end{table}

\subsection{A general summary of the results}

\label{subsec:general}

In order to get a summarized view of the evaluated heuristics, we
present six figures. In these figures we have included
cpu time and number of visited nodes for the three major
conflict-driven variants (\emph{dom/wdeg}, ``\emph{alldell}'' and ``\emph{fully assigned}'')
and we have compared them graphically to the Impact
heuristic (which has the best performance among the impact based
heuristics).

\begin{figure}
  \begin{tabular}{cc}
    \includegraphics[height=2.2in]{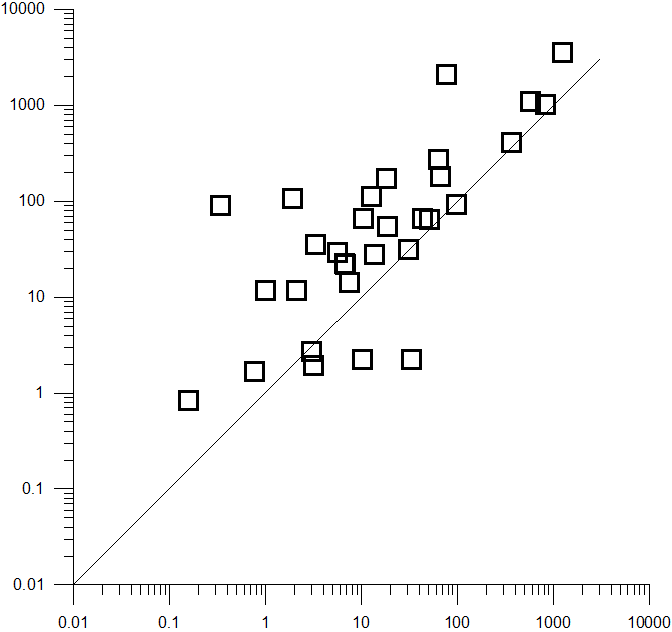} &
    \includegraphics[height=2.2in]{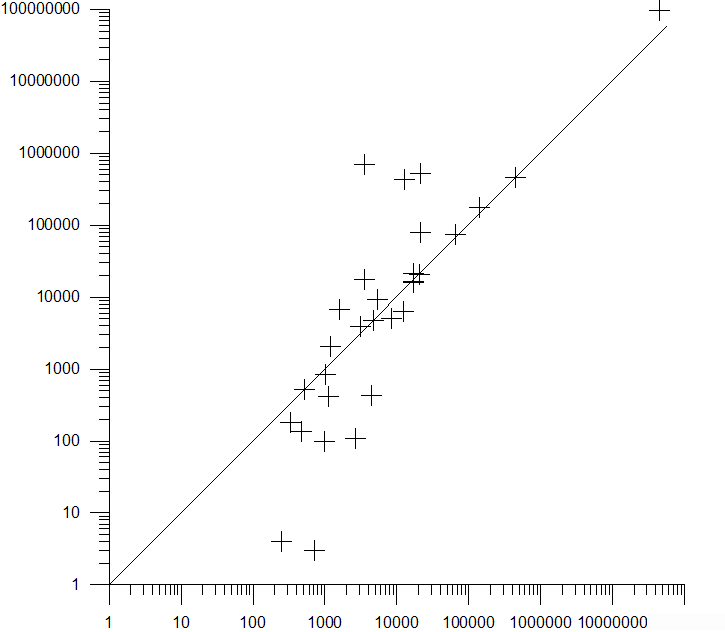}\\
    (a)&(b)\\
    \includegraphics[height=2.2in]{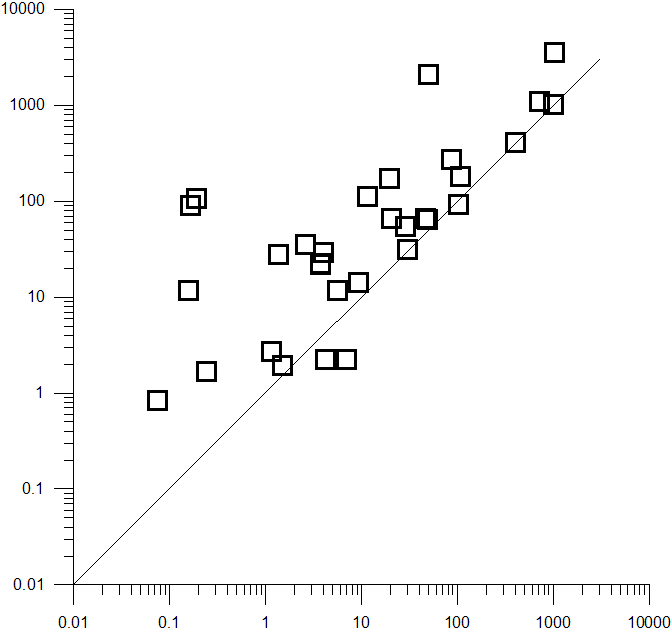} &
    \includegraphics[height=2.2in]{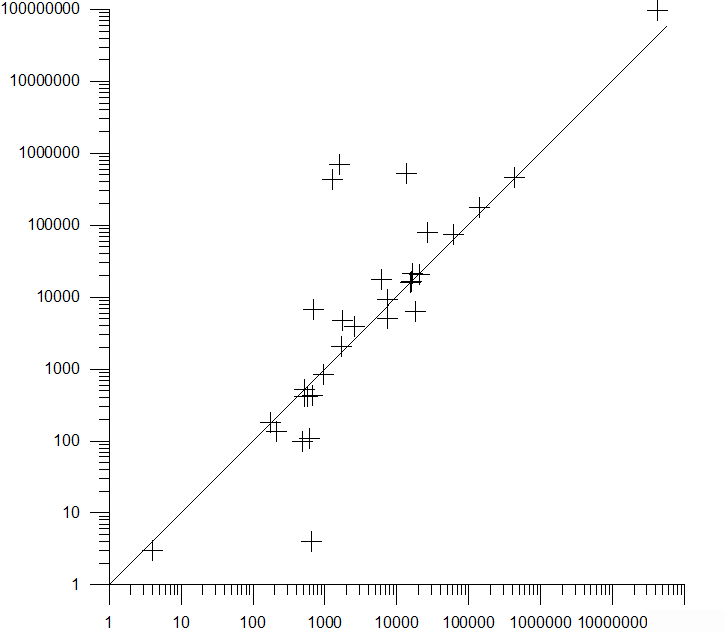}\\
    (c)&(d)\\
    \includegraphics[height=2.2in]{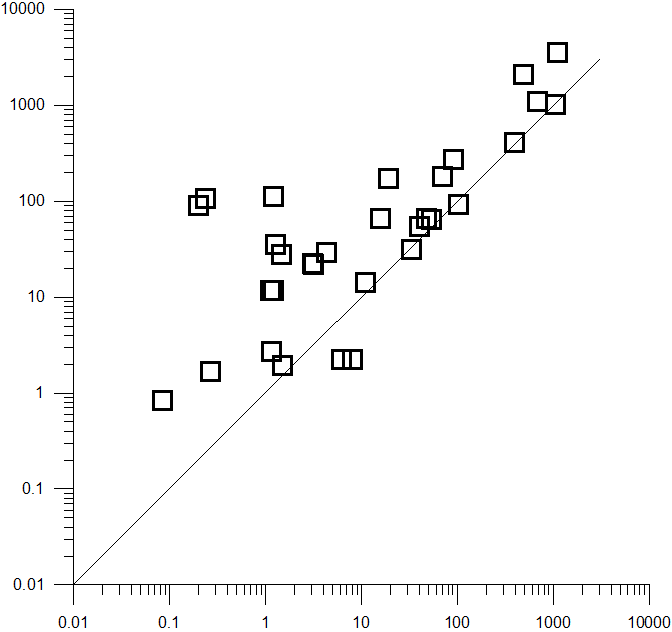} &
    \includegraphics[height=2.2in]{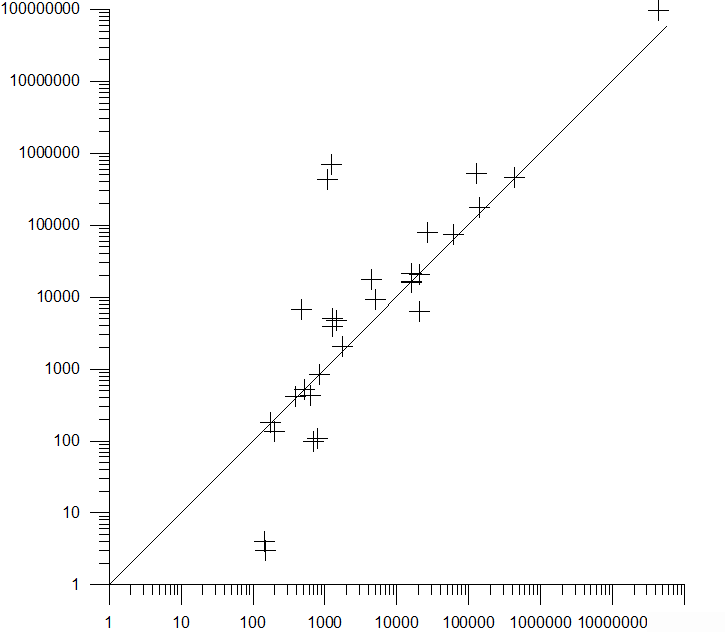}\\
    (e)&(f)
  \end{tabular}
\caption{A summary view of run times (left figures) and nodes
visited (right figures), for $dom/wdeg$ and $impact$ heuristics
(figures (a),(b)), ``\emph{alldell}'' and $impact$ heuristics (figures (c),(d)),
``\emph{fully assigned}'' and $impact$ heuristics (figures (e),(f)).
}
\label{fig:plot}
\end{figure}

Results are collected in Figure~\ref{fig:plot}. The left plots in
these figures correspond to cpu times and the right plots to visited
nodes. Each point in these plots, shows the cpu time (or nodes
visited) for one instance from all the presented benchmarks. The
$y$-axes represent the solving time (or nodes visited) for the
Impact heuristic and the $x$-axes the corresponding values for the
$dom/wdeg$ heuristic (Figures (a) and (b)), ``\emph{alldell}'' heuristic (Figures (c) and (d)) and
``\emph{fully assigned}'' heuristic (Figures (e) and (f)). Therefore, a point above line $y=x$
represents an instance which is solved faster (or with less node
visits) using one of the conflict-driven heuristics. Both axes are
logarithmic.

As we can clearly see from Figure~\ref{fig:plot} (left plots),
conflict-driven heuristics are almost always faster. Concerning the
numbers of visited nodes, the right plots do not reflect an identical
performance. Although it seems that in most cases conflict-driven
heuristics are making a better exploration in the search tree,
there is a considerable set of instances where the Impact
heuristic visit less nodes.

The main reason for this variation in performance (cpu time versus nodes visited)
that the impact heuristic has, is the time consuming process of initialization.
The idea of detecting choices which are responsible for the strongest domain reduction
is quite good. This is verified by the left plots of Figure~\ref{fig:plot}.
But the additional computational overhead of computing the ``best'' choices, really affect
the overall performance of the impact heuristic (Figure~\ref{fig:plot}, right plots).
As our experiments showed the impact heuristic cannot handle efficiently problems
which include variables with relatively large domains. For example in the RLFA problems
where we have 680 variables with at most 44 values in their domains results in Table~\ref{table:rlfap}
verified our hypothesis. On the other hand in problems where variables have only a few values in their domains
(as in the Boolean instances of Section~\ref{subsec:bool}) results showed that
the impact heuristic is quite competitive.

Finally, it has to be noted that the dominant conflict-driven
heuristics are generic and can be also applied in solvers that use
2-way branching and make heavy use of propagators\footnote{A
propagator is essentially a specialized filtering algorithm for a
constraint.} for global constraints, as do most commercial
solvers. In the case of 2-way branching the heuristics can be
applied in exactly the same way as in d-way branching. In the case
of global constraints simple modifications may be necessary, for
example to associate each constraint with a weight independent from
the propagator chosen for the constraint. But having said these,
it remains to be verified experimentally whether the presence of
global constraints or the application of 2-way instead of d-way
branching influence the relevant performance of the heuristics.

\section{Conflict-driven revision ordering heuristics}

\label{sec:revision}

Having demonstrated that conflict-driven heuristics such as
\emph{dom/wdeg} are the dominant modern variable ordering
heuristics, we turn our attention to the use of failures
discovered during search in a different context. To be precise, we
investigate their use in devising heuristics for the ordering of
the (G)AC revision list.

It is well known that the order in which the elements of the
revision list are processed affects the overall cost of the search
\cite{wallace92,boussem04,schulte08}. This is true for solvers
that implement variable or constraint based propagation as well as
for propagator oriented solvers like Ilog Solver and Geocode. In
general, revision ordering and variable ordering heuristics have
different tasks to perform when used by a search algorithm like
MAC. Prior to the emergence of conflict-driven heuristics there
was no way to achieve an interaction with each other, i.e. the
order in which the revision list was organized during the
application of AC could not affect the decision of which variable
to select next (and vice versa). The contribution of revision
ordering heuristics to the solver's efficiency was limited to the
reduction of list operations and constraint checks.

In this section we first show that the ordering of the revision
list can affect the decisions taken by a conflict-driven DVO
heuristic. That is, different orderings can lead to different
parts of the search space being explored. Based on this
observation, we then present a set of new revision ordering
heuristics that use constraint weights, which can not only reduce
the numbers of constraints checks and list operations, but also
cut down the size of the explored search tree. Finally, we
demonstrate that some conflict-driven DVO heuristics, e.g.
``\emph{alldel}'' and ``\emph{fully assigned}'', are less amenable
to changes in the revision list ordering than others (e.g.
\emph{dom/wdeg}).

First of all, to illustrate the interaction between a
conflict-driven variable ordering heuristic and revision list
orderings, we give the following example.

\begin{example}
Assume that we want to solve a CSP $(X,D,C)$, where \emph{X} contains \emph{n} variables \{\emph{$x_1, x_2,..., x_n$}\},
using a conflict-driven variable ordering
heuristic (e.g. dom/wdeg), and that at some point during search
and propagation the variables pending for revision are $x_1$ and
$x_5$. Also assume that two of the constraints in the problem are
$x_1>x_2$ and $x_5>x_6$, and that the domains of $x_1,x_2,x_5,x_6$
are as follows: $D(x_1)=D(x_5)=\{0,1\}$, $D(x_2)=D(x_6)=\{2,3\}$.
Given these constraints and domains, the revision of $x_1$ against
$x_2$ would result in the DWO of $x_1$, and the revision of $x_5$
against $x_6$ would result in the DWO of $x_5$. Independent of
which variable is selected to be revised first (i.e. either $x_1$
or $x_5$), a DWO will be detected and the solver will reject the
current variable assignment. However, depending on the order of
revisions, the \emph{dom/wdeg} heuristic will increase the weight
of a different constraint. To be precise, if a revision ordering
heuristic $R_1$ selects to revise $x_1$ first then the DWO of
$D(x_1)$ will be detected and the weight of constraint $c_{12}$
will be increased by 1. If some other revision ordering heuristic
$R_2$ selects $x_5$ first then the DWO of $D(x_5)$ will be
detected, but this time the weight of constraint $c_{56}$ will be
increased by 1. Since increases in constraint weights affect the
subsequent choices of the variable ordering heuristic, $R_1$ and
$R_2$ can lead to different future decisions for variable
instantiation. Thus, $R_1$ and $R_2$ may guide search to different
parts of the search space. \label{example:revision}
\end{example}

From the above example it becomes clear that the revision ordering
can have an important impact on the performance of conflict-driven
heuristics like \emph{dom/wdeg}. One might argue that a way to
overcome this is to continue propagation after the first DWO is
detected, try to identify all possible DWOs and increase the
weights of all constraints involved in failures. The problem with
this approach is threefold: First, it may increase the cost of
constraint propagation significantly, second it requires
modifications in the way all solvers implement constraint
propagation (i.e. stopping after a failure is detected), and third,
experiments we have run showed that the possibility of more than one
DWO occurring is typically very low. As we
will discuss in Section~\ref{section:dependency}, some variants of
\emph{dom/wdeg} are less amenable to different revision orderings,
i.e. their performance do not depend on the ordering as much,
without having to implement this potentially complex approach.

In the following we first review three standard implementations of
revision lists for AC, i.e. the arc-oriented, variable-oriented,
and constraint-oriented variants. Then, we summarize the major
revision ordering heuristics that have been proposed so far in
the literature, before describing the new efficient revision
ordering heuristics we propose.

\subsection{AC variants}

The numerous AC algorithms that have been proposed can be
classified into {\em coarse grained} and {\em fine grained}.
Typically, coarse grained algorithms like AC-3 \cite{mack77} and
its extensions (e.g. AC2001/3.1 \cite{bryz05} and AC-$3_d$
\cite{dongen02}) apply successive revisions of arcs, variables, or
constraints. On the other hand, fine grained algorithms like
AC-4 \cite{mh86} and AC-7 \cite{bfr95} use various data
structures to apply successive revisions of
variable-value-constraint triplets. Here we are concerned with
coarse grained algorithms, and specifically AC-3. There are two
reasons for this. First, although AC-3 does not have an optimal
worst-case time complexity, as the fine grained algorithms do, it
is competitive and often better in practice and has the additional
advantage of being easy to implement. 
Second, many constraint solvers that
can handle constraints of any arity follow the philosophy of
coarse grained AC algorithms in their implementation of constraint
propagation. That is, they apply successive revisions of variables
or constraints. Hence, the revision ordering heuristics we
describe below can be easily incorporated into most of the
existing solvers.

As mentioned, the AC-3 algorithm can be implemented using a
variety of propagation schemes. We recall here the three variants,
following the presentation of \cite{boussem04}, which respectively
correspond to algorithms with an arc-oriented, variable-oriented
or constraint-oriented propagation scheme.

The first one (arc-oriented propagation) is the most commonly
presented and used because of its simple and natural structure.
Algorithm~\ref{alg:arc} depicts the main procedure. As explained,
an arc is a  pair
(\emph{$c_{ij},x_j$}) which corresponds to a directed constraint. 
Hence, for each binary constraint $c_{ij}$ involving variables
$x_i$ and $x_j$ there are two arcs, (\emph{$c_{ij},x_j$}) and
(\emph{$c_{ij},x_i$}). Initially, the algorithm inserts all arcs in
the revision list \emph{Q}. Then, each arc (\emph{$c_{ij},x_j$}) is
removed from the list and revised in turn. If any value in
\emph{$D(x_j)$} is removed when revising (\emph{$c_{ij},x_j$}), all
arcs pointing to \emph{$x_j$} (i.e. having $x_i$ as second element
in the pair), except (\emph{$c_{ij},x_i$}), will be inserted in
\emph{Q} (if not already there) to be revised.
Algorithm~\ref{alg:revise} depicts function
\emph{REVISE}(\emph{$c_{ij},x_j$}) which seeks supports for the
values of $x_j$ in $D(x_i)$. It removes those values in
\emph{$D(x_j)$} that do not have any support in \emph{$D(x_i)$}.
The algorithm terminates when the list \emph{Q} becomes empty.


\newcommand{\arc}{\ensuremath{\mbox{\sc arc-oriented AC3}}}
\begin{algorithm}[h!]
\caption{$\arc$}\label{alg:arc}
\begin{algorithmic}[1]
\STATE $Q\leftarrow$\{(\emph{$c_{ij}$, $x_j$}) $|$ \emph{$c_{ij}
\in C$} and \emph{$x_{j} \in vars(c_{ij})$}\}
\WHILE{$Q \neq \emptyset$}
  \STATE select and delete an arc (\emph{$c_{ij}$, $x_j$}) from \emph{Q}
  \IF{REVISE(\emph{$c_{ij}$, $x_j$})}
     \STATE \emph{Q} $\leftarrow$ \emph{Q} $\cup$ \{(\emph{$c_{kj}$, $x_k$}) $|$ \emph{$c_{kj} \in C, k \neq i$}\}
  \ENDIF
\ENDWHILE
\end{algorithmic}
\end{algorithm}

\newcommand{\revise}{\ensuremath{\mbox{\sc revise}}}
\begin{algorithm}[h!]
\caption{$\revise(\emph{$c_{ij}$, $x_i$})$}\label{alg:revise}
\begin{algorithmic}[1]
\STATE DELETE $\leftarrow$ \FALSE \FOR{each \emph{a} $\in$
\emph{$D(x_i)$}}
   \IF{$\nexists$ \emph{b} $\in$ \emph{$D(x_j)$} such that \emph{(a, b)} satisfies $c_{ij}$}
     \STATE delete \emph{a} from \emph{$D(x_i)$}
     \STATE DELETE $\leftarrow$ \TRUE
   \ENDIF
\ENDFOR \RETURN DELETE
\end{algorithmic}
\end{algorithm}

\newcommand{\varOrient}{\ensuremath{\mbox{\sc variable-oriented AC3}}}
\begin{algorithm}[h!]
\caption{$\varOrient$}\label{alg:varOrient}
\begin{algorithmic}[1]
\STATE \emph{Q} $\leftarrow$ \{$x_i$ $|$ $x_i \in X$\} \STATE
$\forall$ $c_{ij} \in C, \forall x_i \in$ \emph{vars}($c_{ij}$),
\emph{ctr}($c_{ij}, x_i$) $\leftarrow$ 1 \WHILE{\emph{Q} $\neq$
$\emptyset$}
  \STATE get $x_i$ from \emph{Q}
  \FOR{each $c_{ij}$ $|$ $x_i \in$ \emph{vars}($c_{ij}$)}
     \STATE \textbf{if} \emph{ctr}($c_{ij}, x_i$) = 0  \textbf{then continue}
     \FOR{each $x_j \in vars(c_{ij})$}
        \STATE \textbf{if} \emph{NEEDS-NOT-BE-REVISED}($c_{ij}, x_j$) \textbf{then continue}
          \STATE \emph{nbRemovals} $\leftarrow$ \emph{REVISE}($c_{ij}, x_j$)
          \IF{\emph{nbRemovals} $>$ 0}
            \STATE \textbf{if} \emph{dom}($x_j$) = $\emptyset$ \textbf{then return} \FALSE
            \STATE \emph{Q} $\leftarrow$ \emph{Q} $\cup$ \{$x_j$\}
            \FOR{each $c_{jk}$ $|$ $c_{jk} \neq c_{ij} \wedge x_j \in$ \emph{vars}($c_{jk}$)}
               \STATE \emph{ctr}($c_{jk}, x_j$) $\leftarrow$ \emph{ctr}($c_{jk}, x_j$) + \emph{nbRemovals}
            \ENDFOR
          \ENDIF
     \ENDFOR
     \STATE \textbf{for} each $x_j \in$ \emph{vars}($c_{ij}$) \textbf{do} \emph{ctr}($c_{ij}, x_j$) $\leftarrow$ 0
  \ENDFOR
\ENDWHILE \RETURN \TRUE
\end{algorithmic}
\end{algorithm}

\newcommand{\neednot}{\ensuremath{\mbox{\sc needs-not-be-revised($c_{ij}, x_i$)}}}
\begin{algorithm}[h!]
\caption{$\neednot$}\label{alg:neednot}
\begin{algorithmic}[1]
\STATE \textbf{return} (\emph{ctr}($c_{ij}, x_i$) $>$ 0 and $\nexists x_j
\in$ \emph{vars}($c_{ij}$) $|$ $x_j \neq x_i$ $\wedge$
\emph{ctr}($c_{ij}, x_j$) $>$ 0)
\end{algorithmic}
\end{algorithm}

\newcommand{\conOrient}{\ensuremath{\mbox{\sc constraint-oriented AC3}}}
\begin{algorithm}[h!]
\caption{$\conOrient$}\label{alg:conOrient}
\begin{algorithmic}[1]
\STATE \emph{Q} $\leftarrow$ \{$c_{ij}$ $|$ $c_{ij} \in C$\}
\STATE $\forall$ $c_{ij} \in C, \forall x_i \in$
\emph{vars}($c_{ij}$), \emph{ctr}($c_{ij}, x_i$) $\leftarrow$ 1
\WHILE{\emph{Q} $\neq$ $\emptyset$}
  \STATE get $c_{ij}$ from \emph{Q}
     \FOR{each $x_j \in vars(c_{ij})$}
        \STATE \textbf{if} \emph{NEEDS-NOT-BE-REVISED}($c_{ij}, x_j$) \textbf{then continue}
          \STATE \emph{nbRemovals} $\leftarrow$ \emph{REVISE}($c_{ij}, x_j$)
          \IF{\emph{nbRemovals} $>$ 0}
            \STATE \textbf{if} \emph{dom}($x_j$) = $\emptyset$ \textbf{then return} \FALSE
            \FOR{each $c_{jk}$ $|$ $c_{jk} \neq c_{ij} \wedge x_j \in$ \emph{vars}($c_{jk}$)}
               \STATE \emph{Q} $\leftarrow$ \emph{Q} $\cup$ \{$x_j$\}
               \STATE \emph{ctr}($c_{jk}, x_j$) $\leftarrow$ \emph{ctr}($c_{jk}, x_j$) + \emph{nbRemovals}
            \ENDFOR
          \ENDIF
     \ENDFOR
     \STATE \textbf{for} each $x_j \in$ \emph{vars}($c_{ij}$) \textbf{do} \emph{ctr}($c_{ij}, x_j$) $\leftarrow$ 0
\ENDWHILE \RETURN \TRUE
\end{algorithmic}
\end{algorithm}


The variable-oriented propagation scheme was proposed by McGregor
\cite{mcgreg79} and later studied in \cite{chmeiss98}. Instead of
keeping arcs in the revision list, this variant of AC-3 keeps
variables. The main procedure is depicted in
Algorithm~\ref{alg:varOrient}. Initially, all variables are
inserted in the revision list \emph{Q}. Then each variable $x_i$
is removed from the list and each constraint involving $x_i$ is
processed. For each such constraint $c_{ij}$ we revise the arc
($x_j$,$x_i$). If the revision removes some values from the domain
of $x_j$, then variable $x_j$ is inserted in \emph{Q} (if not
already there).

Function \emph{NEEDS-NOT-BE-REVISED} given in
Algorithm~\ref{alg:neednot}, is used to determine relevant
revisions. This is done by associating a counter
$ctr$($c_{ij}$,$x_i$) with any arc ($x_i$,$x_j$). The value of the
counter denotes the number of removed values in the domain of
variable $x_i$ since the last revision involving constraint
$c_{ij}$. If $x_i$ is the only variable in $vars(c_{ij})$ that has
a counter value greater than zero, then we only need to revise arc
($x_j$,$x_i$). Otherwise, both arcs are revised.

The constraint-oriented propagation scheme is depicted in
Algorithm~\ref{alg:conOrient}. This algorithm is an analogue to
Algorithm~\ref{alg:varOrient}. Initially, all constraints  are
inserted in the revision list \emph{Q}. Then each constraint
$c_{ij}$ is removed from the list and each variable $x_j \in
vars(c_{ij})$ is selected and revised. If the revision of the
selected arc ($c_{ij}$, $x_j$) is fruitful, then the reinsertion
of the constraint $c_{ij}$ in the list is needed. As in the
variable-oriented scheme, the same counters are also used here to
avoid useless revisions.

\subsection{Overview of revision ordering heuristics}

Revision ordering heuristics is a topic that has received
considerable attention in the literature. The first systematic
study on this topic was carried out by Wallace and Freuder, who
proposed a number of different heuristics that
can be used with the arc-oriented variant of AC-3
\cite{wallace92}. These heuristics, which are defined for binary
constraints, are based on three major features of CSPs: (i) the
number of acceptable pairs in each constraint (the constraint size
or satisfiability), (ii) the number of values in each domain and
(iii) the number of binary constraints that each variable
participates in (the degree of the variable). Based on these
features, they proposed three revision ordering heuristics: (i)
ordering the list of arcs by increasing relative satisfiability
(\emph{sat up}), (ii) ordering by increasing size of the domain of
the variables (\emph{dom j up}) and (iii) ordering by descending
degree of each variable (\emph{deg down}).

The heuristic \emph{sat up} counts the number of acceptable pairs
of values in each constraint (i.e the number of tuples in the
Cartesian product built from the current domains of the variables
involved in the constraint) and puts constraints in the list in
ascending order of this count. Although this heuristic reduces the
list additions and constraint checks, it does not speed up the
search process. When a value is deleted from the domain of a
variable, the counter that keeps the number of acceptable arcs has
to be updated. This process is usually time consuming because the
algorithm has to identify the constraints in which the specific
variable participates and to recalculate the counters with
acceptable value pairs. Also an additional overhead is needed to
reorder the list.

The heuristic \emph{dom j up} counts the number of remaining
values in each variable's current domain during search. Variables
are inserted in the list by increasing size of their domains. This
heuristic reduces significantly list additions and constraint
checks and is the most efficient heuristic among those proposed in
\cite{wallace92}.

The \emph{deg down} heuristic counts the current degree of each
variable. The initial \emph{degree} of a variable $x_i$ is the
number of variables that share a constraint with $x_i$. During
search, the \emph{current degree} of $x_i$ is the number of
unassigned variables that share a constraint with $x_i$. The
\emph{deg down} heuristic sorts variables in the list by
decreasing size of their current degree. As noticed in
\cite{wallace92} and confirmed in \cite{boussem04}, the (\emph{deg
down}) heuristic does not offer any improvement.

Gent et al. \cite{gent97} proposed another heuristic called
$k_{ac}$. This heuristic is based on the number of acceptable
pairs of values in each constraint and tries to minimize the
constrainedness of the resulting subproblem. Experiments have
shown that $k_{ac}$ is time expensive but it performs less
constraint checks when compared to \emph{sat up} and \emph{dom j
up}.

Boussemart et al. \cite{boussem04} performed an empirical investigation of the
heuristics of \cite{wallace92} with respect to the different
variants (arc, variable and constraint) of AC-3.
In addition, they introduced some new heuristics. Concerning the
arc-oriented AC-3 variant, they have examined the \emph{dom j up}
as a stand alone heuristic (called \emph{$dom^v$}) or together
with \emph{deg down} which is used in order to break ties (called
$ddeg \circ dom^v$). Moreover, they proposed the ratio \emph{sat
up/dom j up} (called $dom^c / dom^v$) as a new heuristic.
Regarding the variable-oriented variant, they adopted the
\emph{$dom^v$} and $ddeg$ heuristics from \cite{wallace92} and
proposed a new one called $rem^v$. This heuristic corresponds to
the greatest proportion of removed values in a variable's domain.
For the constraint-oriented variant they used $dom^c$ (the
smallest current domain size) and $rem^c$ (the greatest proportion
of removed values in a variable's domain). Experimental results
showed that the variable-oriented AC-3 implementation with the
$dom^v$ revision ordering heuristic (simply denoted $dom$
hereafter) is the most efficient alternative.

\subsection{Revision ordering heuristics based on constraint weights}

\label{subsec:newheuristics}

The heuristics described in the previous subsection, and
especially $dom$, improve the performance of AC-3 (and MAC) when
compared to the classical queue or stack implementation of the
revision list. This improvement in performance is due to
the reduction in list additions and constraint checks. A key
principle that can have a positive effect on the performance of
the AC algorithms is the ``ASAP principle'' by Wallace and Freuder
\cite{wallace92} which urges to ``remove domain values as soon as
possible''. Considering revision ordering heuristics this
principle can be translated as follows: When AC is applied during
search (within an algorithm such as MAC), to reach as early as
possible a failure (\emph{DWO}), order the revision list by
putting first the arc or variable which will guide you to early
value deletions and thus, most likely, earlier to a \emph{DWO}.

To apply the ``ASAP principle'' in revision ordering heuristics,
we must use some metric to compute which arc (or variable) in the
AC revision list is the most likely to cause failure. Until now,
constraint weights have only been used for variable selection. In
the next paragraphs we describe a number of new revision ordering
heuristics for all three AC-3 variants. These heuristics use
information about constraint weights as a metric to order the AC
revision list and they can be used efficiently in
conjunction with conflict-driven variable ordering heuristics to boost search.

The main idea behind these new heuristics is to handle as early as
possible potential \emph{DWO-revisions} by appropriately ordering
the arcs, variables, or constraints in the revision list. In this
way the revision process of AC will be terminated earlier and thus
constraint checks can be reduced significantly. Moreover, with
such a design we may be able to avoid many \emph{redundant
revisions}. As will become clear, all of the proposed heuristics
are lightweight (i.e. cheap to compute) assuming that the weights
of constraints are updated during search.


Arc-oriented heuristics are tailored for the arc-oriented variant
where the list of revisions \emph{Q} stores arcs of the form
($c_{ij}$,$x_i$). Since an arc consists of a constraint $c_{ij}$
and a variable $x_i$, we can use information about the weight of
the constraint, or the weight of the variable, or both, to guide
the heuristic selection. These ideas are the basis of the proposed
heuristics described below. For each heuristic we specify the arc
that it selects. The names of the heuristics are preceded by an
``a'' to denote that they are tailored for arc-oriented
propagation.
\begin{itemize}
  \item $a\_wcon$: selects the arc ($c_{ij}$,$x_i$) such that $c_{ij}$ has the highest weight \emph{wcon} among all
  constraints appearing in an arc in \emph{Q}.
  \item $a\_wdeg$: selects the arc ($c_{ij}$,$x_i$) such that $x_i$ has the highest weighted degree \emph{wdeg} among all
variables appearing in an arc in \emph{Q}.
  \item $a\_dom/wdeg$: selects the arc ($c_{ij}$,$x_i$) such that $x_i$ has the smallest ratio between current domain size
and weighted degree among all variables appearing
 in an arc in \emph{Q}.
  \item $a\_dom/wcon$: selects the arc ($c_{ij}$,$x_i$) having the smallest ratio between the current domain size of
$x_i$ and the weight of $c_{ij}$ among all arcs in \emph{Q}.
\end{itemize}

The call to one of the proposed arc-oriented heuristics can be
attached to line 3 of Algorithm~\ref{alg:arc}. Note that
heuristics $a\_dom/wdeg$ and $a\_dom/wcon$ favor variables with
small domain size hoping that the deletion of their few remaining
values will lead to a DWO. To strictly follow the ``ASAP
principle'' which calls for early value deletions we intend to
evaluate the following heuristics in the future:
\begin{itemize}
  \item $a\_dom/wdeg\_inverse$: selects the arc ($c_{ij}$,$x_i$) such that $x_j$ has the smallest ratio between current domain size
and weighted degree among all variables appearing
 in an arc in \emph{Q}.
  \item $a\_dom/wcon\_inverse$: selects the arc ($c_{ij}$,$x_i$) having the smallest ratio between the current domain size of
$x_j$ and the weight of $c_{ij}$ among all arcs in \emph{Q}.
\end{itemize}

Heuristics $a\_dom/wdeg\_inverse$ and $a\_dom/wcon\_inverse$ favor
revising arcs ($c_{ij}$,$x_i$) such that $x_j$, i.e. the other
variable in constraint $c_{ij}$, has small domain size. This
is because in such cases it is more likely that some values in
$D(x_i)$ will not be supported in $D(x_j)$, and hence will be
deleted.

Variable-oriented heuristics are tailored for the
variable-oriented variant of AC-3 where the list of revisions
\emph{Q} stores variables. For each of the heuristics given below
we specify the variable that it selects. The names of the
heuristics are preceded by an ``v'' to denote that they are
tailored for variable-oriented propagation.

\begin{itemize}
  \item $v\_wdeg$: selects the variable having the highest weighted degree \emph{wdeg} among all variables in \emph{Q}.
  \item $v\_dom/wdeg$: selects the variable having the smallest ratio between current domain size and \emph{wdeg}
    among all variables in \emph{Q}.
\end{itemize}

The call to one of the proposed variable-oriented heuristics can
be attached to line 4 of Algorithm~\ref{alg:varOrient}. After
selecting a variable, the algorithm revises, in some order, the
constraints in which the selected variable participates (line 5).
Our heuristics process these constraints in descending order
according to their corresponding weight.

Finally, the constraint-oriented heuristic $c\_wcon$ selects a
constraint \emph{$c_{ij}$} from the AC revision list having the
highest weight among all constraints in \emph{Q}. The call to this
heuristic can be attached to line 4 of
Algorithm~\ref{alg:conOrient}. One can devise more complex
constraint-oriented heuristics by aggregating the weighted degrees
of the variables involved in a constraint. However, we have not
yet implemented such heuristics.

\subsection{Experiments with revision ordering heuristics}

In this section we experimentally investigate the behavior of the
new revision ordering heuristics proposed above on several classes
of real world, academic and random problems. We only include results for
the two most significant arc consistency variants: arc and
variable oriented. We have excluded the constraint-oriented
variant since this is not as competitive as the other two
\cite{boussem04}.

We compare our heuristics with $dom$, the most efficient
previously proposed revision ordering heuristic. We also include
results from the standard \emph{fifo} implementation of the
revision list which always selects the oldest element in the list
(i.e. the list is implemented as a queue). In our tests we have
used the following measures of performance: cpu time in seconds
(t), number of visited nodes (n), number of constraint checks
(c) and the number of times (r) a revision ordering heuristic has to
select an element in the propagation list $Q$.

Tables~\ref{table:arcrlfaparc} and~\ref{table:arcrlfapvar} show
results from some real-world RLFAP instances. In the arc-oriented
implementation of AC-3 (Table~\ref{table:arcrlfaparc}), heuristics
$a\_wcon$, mainly, and $a\_dom/wcon$, to a less extent, decrease
the number of constraint checks and list revisions compared to
$dom$. However, the decrease is not substantial and is rarely
leads into a decrease in cpu times. The notable speed-up observed
for problem s11-f6 is mainly due to the reduction in the number of
visited nodes offered by the two new heuristics. $a\_wdeg$ and
$a\_dom/wdeg$ are less competitive, indicating that information
about the variables involved in arcs is less important compared to
information about constraints.

The variable-oriented implementation
(Table~\ref{table:arcrlfapvar}) is clearly more efficient than the
arc-oriented one. This confirms the results of \cite{boussem04}.
Concerning this implementation, heuristic $v\_dom/wdeg$ in most
cases is better than $dom$ and $queue$ in all the measured
quantities (number of visited nodes, constraint checks and list
revisions). Importantly, these savings are reflected on notable
cpu time gains making the variable-oriented heuristic
$v\_dom/wdeg$ the overall winner. Results also show that as the
instances becomes harder, the efficiency of $v\_dom/wdeg$ compared
to $dom$ increases. The variable-oriented $v\_wdeg$ heuristic in
most cases is better than $dom$ but is clearly less efficient than
$v\_dom/wdeg$.

\begin{table}
\caption{Cpu times (t), constraint checks (c), number of list revisions (r) and nodes (n) from frequency allocation problems (hard instances) using arc oriented propagation. The s prefix stands for scen instances.
Best cpu time is in bold.} 
\centering 
\begin{scriptsize}
\begin{tabular}{|c|c|c|c|c|c|c|c|}
\hline & & \multicolumn{6}{|c|}{ARC ORIENTED}\\ \hline
Inst. &  & $queue$ & $dom$ & $a\_wcon$ & $a\_wdeg$ & $a\_dom/wdeg$ & $a\_dom/wcon$ \\
\hline 
s11-f9 & t & 18,8 & {\bf 12,8} & 14,6 & 14,8 & 19 & 14,2 \\
 & c & 25,03M & 19,3M & 13,2M & 20,8M & 21M & 16,8M \\
 & r & 1,1M & 910060 & 529228 & 1,04M & 1,01M & 737803 \\
 & n & 1202 & 1153 & 1155 & 1145 & 1148 & 1159 \\\hline
s11-f8 & t & 37,5 & {\bf 20,3} & 22,5 & 21,9 & 28,5 & 23,5 \\
 & c & 46,5M & 29,3M & 19,1M & 30,1M & 32,9M & 27,5M \\
 & r & 1,95M & 1,3M & 748050 & 1,52M & 1,43M & 1,11M \\
 & n & 1982 & 1830 & 1843 & 1876 & 1832 & 1928 \\ \hline
s11-f7 & t & 257,5 & {\bf 146,5} & 170 & 265,2 & 205,8 & 326,2 \\
 & c & 268,4M & 159,4M & 128,5M & 281,4M & 205,1M & 300M \\
 & r & 13,3M & 10,2M & 6,1M & 17,7M & 12,1M & 15M \\
 & n & 17643 & 14734 & 15938 & 20617 & 15318 & 29845 \\ \hline
s11-f6 & t & 568,5 & 465,2 & {\bf 309,4} & 540,4 & 834,9 & 396,4 \\
 & c & 482,3M & 468,2M & 230,8M & 517,2M & 745,4M & 362,7M \\
 & r & 27,5M & 29,7M & 10,4M & 34,9M & 49,5M & 16,6M \\
 & n & 46671 & 50021 & 29057 & 49201 & 68217 & 35860 \\ \hline
s11-f5 & t & 2821 & 2307 & 3064 & 3234 & 2898 & {\bf 2291} \\
 & c & 2,492G & 2,139G & 2,097G & 2,928G & 2,596G & 1,965G \\
 & r & 137,8M & 157M & 116,5M & 215,7M & 172,2M & 103,3M \\
 & n & 212012 & 217407 & 287017 & 258261 & 185991 & 187363 \\ \hline
 s11-f4 & t & 11216 & {\bf 7774} & 8256 & 10386 & 12520 & 10473 \\
 & c & 9,938G & 7,054G & 5,298G & 9,020G & 10,711G & 8,598G \\
 & r & 533,4M & 523,1M & 311,7M & 681,2M & 738,1M & 464,7M \\
 & n & 753592 & 709196 & 762477 & 832892 & 850446 & 786924 \\ \hline
\end{tabular}
\end{scriptsize}
\label{table:arcrlfaparc} 
\end{table}

\begin{table}
\caption{Cpu times (t), constraint checks (c), number of list revisions (r) and nodes (n) from frequency allocation problems (hard instances) using variable oriented propagation. The s prefix stands for scen instances.
Best cpu time is in bold.} 
\centering 
\begin{scriptsize}
\begin{tabular}{|c|c|c|c|c|c|}
\hline & & \multicolumn{4}{|c|}{VARIABLE ORIENTED}\\ \hline
Inst. &  & $queue$ & $dom$ & $v\_wdeg$ & $v\_dom/wdeg$ \\
\hline 
s11-f9 & t & 14,3 & 10,2 & 10,9 & \textbf{9,9}\\
 & c & 22,6M & 11,4M & 12,9M & 11M\\
 & r & 28978 & 17177 & 20161 & 17048\\
 & n & 1413 & 1117 & 1145 & 1137\\ \hline
s11-f8 & t & 21,2 & 17,3 & 18,5 & \textbf{16,7}\\
 & c & 42,1M & 17,2M & 20,4M & 16,8M\\
 & r & 48568 & 24885 & 28807 & 24819\\
 & n & 2112 & 1842 & 1830 & 1841\\ \hline
s11-f7 & t & 133,7 & 158,1 & 154,5 & \textbf{108,2}\\
 & c & 193,3M & 116,9M & 157,6M & 82,7M\\
 & r & 313568 & 223094 & 263306 & 156160\\
 & n & 12777 & 18773 & 14570 & 13181\\ \hline
s11-f6 & t & 391,4 & 391 & 434,4 & \textbf{269,5}\\
 & c & 306,2M & 263,2M & 413,6M & 192,6M\\
 & r & 426469 & 509474 & 673935 & 340583\\
 & n & 34714 & 46713 & 41609 & 31538\\ \hline
s11-f5 & t & 2473 & 3255 & 2019 & \textbf{1733}\\
 & c & 2,073G & 2,115G & 1,502G & 1,157G\\
 & r & 3,63M & 4,52M & 2,97M & 2,2M\\
 & n & 223965 & 397590 & 190496 & 199854\\ \hline
 s11-f4 & t & 13969 & 11859 & 9490 & \textbf{6669}\\
 & c & 12,059G & 7,512G & 6,915G & 4,322G\\
 & r & 20,3M & 15,9M & 14M & 8,9M\\
 & n & 1,148M & 1,354M & 939094 & 716427\\ \hline
\end{tabular}
\end{scriptsize}
\label{table:arcrlfapvar} 
\end{table}

In Table~\ref{table:langford} we present results from structured
instances belonging to benchmark classes {\em langford} and {\em
driver}. As the variable-oriented AC-3 variant is more efficient
than the arc-oriented one, we only present results from the
former. Results show that on easy problems all heuristics except
$queue$ are quite competitive. But as the difficulty of the
problem increases, the improvement offered by the $v\_dom/wdeg$
revision heuristic becomes clear. On instance driverlogw-09 we can
see the effect that weight based revision ordering heuristics can
have on search. $v\_dom/wdeg$ cuts down the number of node visits
by more than 5 times resulting in a similar speed-up. It is
interesting that $v\_dom/wdeg$ is considerably more efficient than
$v\_wdeg$ and $dom$, indicating that information about domain size
or weighted degree alone is not sufficient to efficiently order
the revision list.

\begin{table}
\caption{Cpu times (t), constraint checks (c), number of list revisions (r) and nodes (n) from structured
problems using variable oriented propagation. Best cpu time is in bold.}
\centering
\begin{scriptsize}
\begin{tabular}{|c|c|c|c|c|c|}
\hline
Instance &  & $queue$ & $dom$ & $v\_wdeg$ & $v\_dom/wdeg$ \\
\hline 
langford-2-9 & t & 56,5 & 46,9 & 60,3 & \textbf{46,2}\\
 & c & 99,6M & 81,7M & 99,9M & 81,5M \\
 & r & 633113 & 533656 & 741596 & 533261\\
 & n & 48533 & 40228 & 49114 & 40363\\ \hline
langford-2-10 & t & 489,8 & 430,6 & 418,9 & \textbf{340,1}\\
 & c & 336,1M & 283,7M & 275,2M & 197,9M \\
 & r & 5,3M & 4,5M & 4M & 2,9M\\
 & n & 337772 & 280600 & 260343 & 208651\\ \hline
langford-3-11 & t & 695,8 & 648,5 & 843,5 & \textbf{513,5}\\
 & c & 408,6M & 352,7M & 468,8M & 256,7M \\
 & r & 2,3M & 1,9M & 2,9M & 1,6M\\
 & n & 99508 & 68042 & 103863 & 65958\\ \hline
langford-4-10 & t & 81,4 & 57,7 & 99,4 & \textbf{41,2}\\
 & c & 52,3M & 33,2M & 59,6M & 21,7M \\
 & r & 150493 & 99646 & 194952 & 75889\\
 & n & 3852 & 2973 & 5759 & 2661\\ \hline
driverlogw-08c & t & 19,4 & 14,7 & \textbf{14,4} & 14,6\\
 & c & 20,8M & 8,6M & 10,9M & 9M \\
 & r & 86809 & 39063 & 40256 & 38748\\
 & n & 3151 & 3040 & 1960 & 2660\\ \hline
driverlogw-09 & t & 174,6 & 411 & 346,3 & \textbf{70,1}\\
 & c & 151,5M & 251,5M & 203,6M & 39,5M \\
 & r & 521358 & 1,05M & 583686 & 139962\\
 & n & 21220 & 41039 & 31548 & 7457\\ \hline
\end{tabular}
\end{scriptsize}
\label{table:langford} 
\end{table}

Finally, in Table~\ref{table:random} we present results from
random and quasi-random problems. In the geo50-20-d4-75-2 , which
is a quasi-random instance we can see that the proposed heuristics
($v\_wdeg$ and $v\_dom/wdeg$) are one order of magnitude faster
than $dom$. This suggest that the small presence of structure is
this problem results in behavior similar to the behavior observed
in the structured instances of Table~\ref{table:langford}.

On the rest of the instances, which are purely random, there is a
large variance in the results. All heuristics seems to lack
robustness and there is no clear winner. The constraint weight
based heuristics can be faster than $dom$ (instance frb30-15-1),
but they can also be significantly slower (frb30-15-2). In all
cases, the large run time differences in favor of one or another
heuristic are caused by corresponding differences in the size of
the explored search tree, as node visits clearly demonstrate.

A plausible explanation for the diversity in the performance of the
heuristics on pure random problems as opposed to structured ones
is the following. When dealing with structured problems, and
assuming we use the variable-oriented variant of AC-3, a weighted
based heuristic like $v\_dom/wdeg$ will give priority for revision
to variables that are involved in hard subproblems and hence will
carry out DWO-revisions faster. This will in turn increase the
weights of constraints that are involved in such hard subproblems
and thus search will focus on the most important parts of the
search space. Pure random instances that lack structure do not in
general consist of hard local subproblems. Thus, different
decisions on which variables to revise first can lead to different
DWO-revisions being discovered, which in turn can guide search
tree to different parts of the search space with unpredictable
results. Note that for structured problems only very few possible
DWO-revisions are present in the revision list at each point in
time, while for random ones there can be a large number of such
revisions.

\begin{table}
\caption{Cpu times (t), constraint checks (c), number of list revisions (r) and nodes (n)
from random problems using variable oriented propagation. Best cpu time is in bold.}
\centering
\begin{scriptsize}
\begin{tabular}{|c|c|c|c|c|c|}
\hline

Instance &  & $queue$ & $dom$ & $v\_wdeg$ & $v\_dom/wdeg$ \\
\hline 
frb30-15-1 & t & 22,3 & 20,9 & 29,3 & \textbf{14,1}\\
 & c & 16,5M & 11,1M & 16,4M & 7,5M \\
 & r & 105626 & 70924 & 102724 & 46727\\
 & n & 3863 & 3858 & 4138 & 2499\\ \hline
frb30-15-2 & t & 84,9 & \textbf{29,7} & 118,9 & 95\\
 & c & 45,7M & 21,8M & 90M & 68,9M \\
 & r & 311040 & 149119 & 624360 & 472124\\
 & n & 15457 & 7935 & 25148 & 24467\\ \hline
frb35-17-1 & t & 125,8 & 193,7 & \textbf{118} & 250,9\\
 & c & 93,9M & 144M & 89,7M & 180,9M \\
 & r & 533694 & 836462 & 514258 & 1,03M\\
 & n & 18587 & 40698 & 19167 & 50611\\ \hline
rand-2-30-15 & t & 1240 & \textbf{74,4} & 98 & 108,1\\
 & c & 114,5M & 53M & 72,5M & 78,1M \\
 & r & 922251 & 443792 & 602582 & 642665\\
 & n & 28725 & 19846 & 20192 & 28766\\ \hline
geo50-20-d4-75-2 & t & 226,1 & 401,8 & \textbf{34,8} & 39,5\\
 & c & 191,8M & 310,3M & 28,2M & 28,8M \\
 & r & 778758 & 1,3M & 117241 & 124163\\
 & n & 20069 & 60182 & 3735 & 5484\\ \hline
\end{tabular}
\end{scriptsize}
\label{table:random} 
\end{table}

\subsection{Dependency of conflict-driven heuristics on the revision
ordering}

\label{section:dependency}

As we showed in the previous section, \emph{dom/wdeg} is strongly
dependent on the order in which the revision list is constructed
and updated during constraint propagation. Looking at the results
in Tables~\ref{table:arcrlfaparc} --~\ref{table:random}, we can see that there are cases where the
differences in cpu performance between $dom$  and $v\_dom/wdeg$ can be up to 5 times. Hence, when
\emph{dom/wdeg} is used as DVO heuristic, we must carefully select
a good revision ordering using for example one of the heuristics
we have proposed in Section~\ref{subsec:newheuristics}. In
contrast, the conflict-driven DVO heuristics ``\emph{alldel}'' and
``\emph{fully assigned}'' are not as amenable to the selection of
revision ordering. To better illustrate this statement, let us
consider the following example.

\begin{example}\rm

Assume that we want to solve a CSP $(X,D,C)$ with $X$: \{$x_1$,
$x_2$, $x_3$, $x_4$\}, by using two different revision ordering
heuristics $R_1$ (lexicographic ordering) and $R_2$ (reverse
lexicographic ordering). For the revision of each $x_i$ $\in$ $X$,
we assume that the following hypotheses are true: \emph{a}) The
revision of $x_1$ is fruitful and it causes the addition of the
variable $x_3$ in the revision list. \emph{b}) The revision of
$x_2$ is fruitful and it causes the addition of the variable $x_4$
in the revision list. \emph{c}) The revision of $x_4$ is fruitful
and it causes the addition of the variable $x_3$. We also assume
the a DWO can only occur either \emph{d}) in $x_4$ after a
sequential revision of $x_2$ and $x_3$ or \emph{e}) in $x_3$ after
a sequential revision of $x_4$ and $x_1$. Finally, assume that at
some point during search only the variables $x_1$ and $x_2$ have
remained in the AC revision list $Q$, but with different orderings
for $R_1$ and $R_2$. That is, $Q_{R_1}$:\{$x_1$,$x_2$\},
$Q_{R_2}$:\{$x_2$,$x_1$\}. Following all these assumptions (which
can exist commonly in any real world CSP), lets now trace the behavior
of both $R_1$ and $R_2$ during problem solving. Considering the
$Q_{R_1}$ list, the revision of $x_1$ is fruitful and adds $x_3$
in the list (due to hypothesis $a$). Now the revision list changes
to $Q_{R_1}$:\{$x_2$,$x_3$\}. The sequential revision of $x_2$ and
$x_3$ leads to the DWO of $x_4$ (due to hypotheses $b$ and $d$).
Considering the $Q_{R_2}$ list, the revision of $x_2$ is fruitful
and adds $x_4$ in the list (due to hypothesis $b$). Now the
revision list changes to $Q_{R_2}$:\{$x_4$,$x_1$\}. The sequential
revision of $x_4$ and $x_1$ leads to the DWO of $x_3$. (due to
hypotheses $c$ and $e$).

\end{example}

From the above example it is clear that although only one DWO is
identified in the revision list, both $x_1$ and $x_2$ can be
responsible for this. In $R_1$ where $x_1$ is the DWO variable, we
can say that $x_2$ is also a ``potential'' DWO variable i.e. it
would be a DWO variable, if the $R_2$ revision ordering was used.
Although the \emph{dom/wdeg} heuristic ignores all the ``potential''
DWO variables, the other two DVO heuristics,``\emph{alldel}'' and
``\emph{fully assigned}'', take into account their contribution.
The former heuristic increases the weights for every constraint
that causes a value deletion,  and thus succeeds to increase the
weights of the constraints related to the ``potential'' DWO
variables. The latter heuristic increases the weights of
constraints that participate in fruitful revisions (only for
revision lists that lead to a DWO), and thus is able to frequently
identify ``potential'' DWO variables.

To experimentally verify the strong dependance of $dom/wdeg$
heuristic on the revision ordering and the ability of the ``\emph{alldel}''
and ``\emph{fully assigned}'' heuristics to be less dependent, we have
computed the variance in the number of node visits for the three conflict-driven
 heuristics on some selected instances.

The variance is a measure of how spread out a distribution of a variable's values is.
A variable's spread is the degree to which the values of the variable
differ from each other. If all values of the variable were about
equal, the variable would have very little spread.
In other words, it is a measure of variability.
In our case the measured variable $x$ is the number of visited nodes for the conflict-driven heuristics.
For each conflict-driven heuristic the $x$ variable can take $N$=3 values. That is,
the number of visited nodes when any one of the 3 main revision ordering heuristics ($queue$, $dom$,
$v\_dom/wdeg$) is used.

The variance is calculated by taking the arithmetic mean of the squared
differences between each value and the mean value, using the following equation:

\begin{equation}
VARIANCE =  \frac{\sum (x-\bar{x})^2}{N}
\end{equation}

where $x$ is the number of node visits when a specific revision ordering heuristic is used and
$\bar{x}$ is the mean number of visited nodes of the
$N$=3 main revision ordering heuristics ($queue$, $dom$,
$v\_dom/wdeg$).

The smaller the variance of a conflict-driven heuristic, the less
the dependance from the selected revision ordering heuristic.
Results from these experiments are depicted in
Table~\ref{table:devsq}. As we can see, in
almost all cases the $dom/wdeg$ heuristic displays the highest
variance, while the other two conflict-driven heuristics in most
cases have smaller values. This suggests that indeed
the ``\emph{alldel}'' and ``\emph{fully assigned}'' heuristics
are less amenable to changes in the revision ordering than
\emph{dom/wdeg} and therefore can be more robust.

\begin{table}
\caption{The computed variances for the three conflict-driven heuristics. Best values is in bold.}
\centering
\begin{scriptsize}
\begin{tabular}{|c|c|c|c|c|c|}
\hline

Instance & $dom/wdeg$ & $alldel$ & $fully$ $assigned$ \\
\hline 
scen-11 & 96732 & 7432 & \bf{67} \\ \hline
scen-11-f8 & 6893 & 2127 & \bf{701} \\ \hline
scen-11-f7 & 3974589 & 6384509 & \bf{1454538} \\ \hline
jnh01 & 6123 & \bf{80} & 41280 \\ \hline
jnh17 & 1316 & \bf{52} & 91 \\ \hline
jnh201 & 4238 & 12 & \bf{7} \\ \hline
jnh301 & 66738 & 19783 & \bf{91} \\ \hline
langford-2-10 & 7564932 & \bf{4547893} & 10923451 \\ \hline
driverlogw-08c & 291287 & 8465 & \bf{912} \\ \hline
driverlogw-09 & 71643951 & 19821345 & \bf{13189345} \\ \hline
will199GPIA-5 & 1139 & \bf{0} & 3717 \\ \hline
will199GPIA-6 & 5313746 & 860138 & \bf{614930} \\ \hline
\end{tabular}
\end{scriptsize}
\label{table:devsq} 
\end{table}

Finally, it would be interesting to apply similar ideas as the
ones presented in Section~\ref{sec:revision} to propagator-heavy
solvers. Constraint propagation in such solvers is not handled by
a revision list of variables or constraints, but they do use
heuristics to choose the order in which propagators will be
applied \cite{schulte08}. Hence taking exploiting information such
as constraint weights might be beneficial.

\section{Conclusions and future work}

\label{sec:conclusion}

In this paper we experimentally evaluated the most recent and
powerful variable ordering heuristics, and new variants of them,
over a wide range of academic, random and real world problems.
These heuristics can be divided in two main categories: heuristics
that exploit information about failures gathered throughout search
and recorded in the form of constraint weights and heuristics that
measure the importance/impact of variable assignments for reducing
the search space. Results demonstrate that heuristics based on
failures have much better cpu performance. Although impact based
heuristics are in general slow, there are some cases where they
perform a smarter exploration of the search tree resulting in
fewer node visits. Among the tested conflict-driven heuristics,
\emph{dom/wdeg} seems to be the faster followed closely by its
variants ``\emph{alldel}'' and ``\emph{fully assigned}''.

We also showed how information about failures can be exploited to
design efficient revision ordering heuristics for algorithms that
maintain (G)AC using coarse grained arc consistency algorithms.
The proposed heuristics order the revision list by trying to carry
out possible DWO-revisions as soon as possible. Importantly, these
heuristics can not only reduce the numbers of constraint checks
and list operations but they can also have a significant effect on
search. Among the revision ordering heuristic we experimented
with, the one with best performance was $v\_dom/wdeg$ in the
variable-oriented implementation of arc consistency.

Finally, we experimentally demonstrated that although \emph{dom/wdeg} is the most
efficient conflict-driven heuristic, other conflict-driven
heuristics like `\emph{alldel}'' and ``\emph{fully assigned}''
have the advantage of being less dependent on the revision
ordering heuristic used. Hence, the performance of \emph{dom/wdeg}
can be less predictable under different revision orderings.

As a future work, we intent to experimentally examine the behavior
of the modern variable ordering heuristics, on problems with global constraints.
Concerning revision ordering heuristics, we plan to evaluate the inverse arc-oriented
heuristics: $a\_dom/wdeg\_inverse$ and $a\_dom/wcon\_inverse$, which favor
revising arcs ($c_{ij}$,$x_i$) such that $x_j$, has small domain size.

\section*{Acknowledgements}

We would like to thank the anonymous reviewers of an earlier
version of this paper for their constructive comments that helped
improve the quality of the paper.

\bibliography{extra}

\end{document}